\documentclass[11pt]{article}
\usepackage{fancyhdr}
\usepackage{amssymb}
\usepackage{amsmath}
\usepackage{mathrsfs}
\usepackage{color}
\usepackage[toc,page]{appendix}
\usepackage{url}
\usepackage{marginnote}
 
\usepackage{graphicx}
\graphicspath{{Figs/}}
\DeclareGraphicsExtensions{.eps,.pdf,.jpg,.gif}

\usepackage{algorithm}
\usepackage{algorithmic}

\def\here{
 \[\vcenter{
\medskip
\hbox to\hsize{\hfill\footnotesize here $\downarrow$}
\smallskip
\hrule
\medskip}\]}

\font\fiverm=cmr5
\def\R{{\bf R}}
\def\N{{\bf N}}
\def\feat{C\mskip -10mu\lower-2pt\hbox{\fiverm 1}\,}
\def\Feat#1{C\mskip -10mu\lower-2pt\hbox{\fiverm #1}\,}
\def\dts{\mathinner{\ldotp\ldotp}} 
\def\dash---{\thinspace---\hskip.16667em\relax} 
\let\del=\partial
\usepackage{epsfig}

\title{\bf Convolutional Networks in \\
Visual Environments
}
\author{Alessandro Betti and Marco Gori \\
SAILab, University of Siena \\
}

\begin{document}

\thispagestyle{empty}

\maketitle

\begin{abstract}
	The puzzle of computer vision might find new challenging
	solutions when we realize that most successful methods are
	working at image level, which is remarkably more difficult
	than processing directly visual streams.  In this paper, we claim
	that their processing naturally leads to formulate the motion
	invariance principle, which enables the construction of a new
	theory of learning with convolutional networks.  The theory
	addresses a number of intriguing questions that arise in
	natural vision, and offers a well-posed computational scheme
	for the discovery of convolutional filters over the
	retina. They are driven by differential equations
	derived from the principle of least cognitive action.
	Unlike traditional convolutional networks, which need massive
	supervision, the proposed theory offers a truly new scenario in
	which feature learning takes place by unsupervised processing 
	of  video signals.
	It is pointed out that an opportune blurring of the video, along
	the interleaving of segments of null signal, make it possible to
	conceive a novel learning mechanism that yields the minimum
	of  the cognitive action.  Basically, while the theory enables 
	the implementation of novel computer vision systems, 
	it is also provides an intriguing explanation of the solution
	that evolution has discovered for humans, where it looks like that
	the video blurring in newborns and the day-night rhythm seem 
	to emerge in a general computational framework, regardless
	of biology.
	%
	%
\end{abstract}

\section{Introduction}
While the emphasis on a general theory of vision was already 
the main objective at the dawn of the discipline~\cite{Marr82}, it has evolved without a systematic exploration of foundations in machine learning. When the target is moved to unrestricted visual environments and the emphasis is shifted from huge labelled databases to a human-like protocol of interaction, we need to go beyond the current peaceful interlude that we are experimenting in vision and machine learning. A fundamental question a good theory is expected to answer is why children can learn to recognize objects and actions from a few supervised examples, whereas nowadays supervised learning approaches strive to achieve this task. In particular, why are they so thirsty for supervised examples? Interestingly, this fundamental difference seems to be deeply rooted in the different communication protocol at the basis of the acquisition of visual skills in children and machines. 

So far, the semantic labeling of pixels of a given video stream has been mostly carried out at frame level. This seems to be the natural outcome of well-established pattern recognition methods working on images, which have given rise to nowadays emphasis on collecting big labelled image databases (e.g.~\cite{imagenet_cvpr09}) 
with the purpose of devising and testing challenging machine learning algorithms. While this framework is the one in which most of nowadays state of the art object recognition approaches have been developing, we argue that there are strong arguments to start exploring the more natural visual interaction that animals experiment in their own environment. 

\marginnote{Learning in wild visual environments.}
This leads to process video instead of image collection, that naturally leads to a paradigm-shift
in the associated processes of learning to see. The idea of shifting to video is very much 
related to the growing interest of {\em learning in the wild} that has been explored in the
last few years (see. e.g.~\url{https://sites.google.com/site/wildml2017icml/}).

A crucial problem that has been recognized by Poggio and Anselmi~\cite{Poggio:2016:VCD} is the need to incorporate visual invariances
into deep nets that go beyond simple translation invariance that is currently characterizing
convolutional networks. They propose an elegant mathematical framework 
on visual invariance and enlighten some intriguing neurobiological connections. 
Overall, the ambition of extracting distinctive features from vision poses a challenging task. 
While we are typically concerned with feature extraction that is independent of classic geometric transformation, it looks like we are still missing the fantastic human skill of capturing distinctive features to recognize ironed and rumpled shirts! There is no apparent difficulty to recognize shirts by keeping the recognition coherence in case we roll up the sleeves, or we simply curl them up into a ball for the laundry basket. Of course, there are neither rigid transformations, like translations and rotation, nor scale maps, that transforms an ironed shirt into the same shirt thrown into the laundry basket. 
Is there any natural invariance?

In this paper, we claim that
motion invariance is in fact the only invariance that we need.
\marginnote{The paradigm-shift of motion invariance}
Translation and scale invariance, that have been the subject of many studies, are in
fact examples of invariances that can be fully gained whenever we develop the 
ability to detect features that are invariant under motion. 
If my inch moves closer and closer to my eyes then any of its representing features 
that is motion invariant will also be scale invariant. The
finger will become bigger and bigger as it approaches my face, but it is still my inch!
Clearly, translation, rotation, and complex deformation invariances  derive from
motion invariance. Humans life always experiments motion, so as the gained visual
invariances naturally arise from motion invariance. Animals with foveal eyes also
move quickly the focus of attention when looking at fixed objects, which means that they
continually experiment motion. Hence, also in case of fixed images, 
conjugate, vergence, saccadic, smooth pursuit, and vestibulo-ocular movements lead to acquire visual information from relative motion.  We claim that the 
production of such a continuous visual stream naturally drives
feature extraction,
since the corresponding convolutional filters are expected not to change during motion. 
The enforcement of this consistency condition creates a mine of visual data
during animal life. Interestingly, the same can happen for machines. 
Of course, we need to compute the optical flow at pixel level so as to enforce
the consistency of all the extracted features. Early studies 
on this problem~\cite{HornAI1981}, along with recent related 
improvements (see e.g.~\cite{Baker:2011})
suggests to determine the velocity field by enforcing brightness invariance. 
As the optical flow is gained, it is used to enforce motion consistency on the
visual features. Interestingly, the theory we propose is quite related to the
variational approach that is used to determine the optical flow in~\cite{HornAI1981}.
It is worth mentioning that an effective visual system must also develop 
features that do not follow motion invariance. These kind of features can be 
conveniently combined with those that are discussed in this paper with the
purpose of carrying out high level visual tasks.

The convolutional filters are somewhat inspired 
from the research activity reported in~\cite{DBLP:journals/cviu/GoriLMM16},
where the authors propose the extraction of visual features as a
constraint satisfaction problem, mostly based on information-based principles
and  early ideas on motion invariance. 

\marginnote{Learning as the minimization of the cognitive action.}
In this paper, the importance of motion invariance is stressed and, moreover,
the solution is derived in the framework of the principle of 
cognitive action~\cite{DBLP:journals/tcs/BettiG16}, which gives rise  to a
time-variant differential equation, where the Lagrangian coordinates
corresponds with the values of the convolutional filters. 
It is pointed out that, under causality conditions, the well-position
of the problems arises thanks to the process of video-blurring taking
place at the beginning of learning, which has also been experimented
in children. 
The learning process can be interpreted in the framework of the minimization
of the cognitive action that offers a self-consistent framework. In particular, if the video signal is almost periodic~\cite{besicovitch1954almost},
then the computational model reduces to an asymptotically stable 
differential equation that yields a sort of statistical consistency.


\section{Driving principles and main results}
\label{driving-principles}
%
We are given a retina $X$, which can 
formally be regarded as a compact subset of the plane; 
for the moment we will not assume any specific 
shape\dash---any deformation of the closed disk will serve.
The purpose of this paper is that of analyzing the mechanisms that give rise
to the construction of local features for any pixel $x \in X$ of the retina, at any
time $t$. These features, along with the video itself, can be regarded as 
visual fields, that are defined on the retina and on a given  horizon of
time  $[0\dts T]$; clearly
the analysis of on-line learning of visual features leads to regard 
the horizon as $[0\dts \infty)$. 
As it will be clear in the remainder of the paper, a set of symbols are extracted
at any layer of a deep architecture, so as any pixel\dash---along with its context\dash---
turns out to be represented by the list of symbols extracted at each layer.
The computational process that we define involves the video
as well as appropriate vector fields that are used to express
a set of pixel-based features properly used to capture contextual
information.  The video, as well as all the involved
fields, are defined on the domain $D=X\times [0\dts T]$.
In what follows, points on the retina will be represented
with two dimensional vectors $x=(x_1,x_2)$ on a defined 
coordinate system on the retina.
The temporal coordinate is usually denoted by
$t$, and, therefore, the video signal on the pair $(x,t)$ is $C(x,t)$. For
further convenience we also define the map $C_t\colon X\to \R^m$ so that
$C_t(x)\equiv C(x,t)$.
The color field can be thought of as a special field that is characterized by the
RGB color components of any single pixel; in this case $m=3$. 
\begin{figure}
		\centering
		\includegraphics{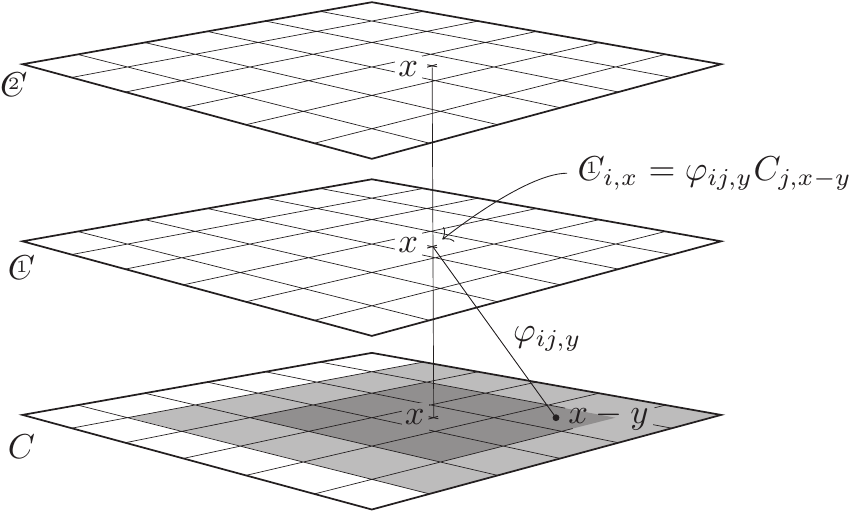}
		\caption{Convolutional computation in a deep network.
		The input is processed by convolutional filters which 
		transform $C \to \feat \to \Feat2$.
                Notice that the features
		are extracted at different level on the same pixel $x$.}
	\label{ConvFig}
\end{figure}

Now, we are concerned with the problem of extracting visual features that, unlike
the components of the video, express the information associated with 
the pair $(x,t)$ and its spatial context. Basically, one would like to extract
visual features that characterize the information in the neighborhood of pixel $x$.
\marginnote{Kernel-based computation feature extraction.}
A possible way of constructing this kind of features is to define\footnote{Throughout the
paper we use the Einstein convention to simplify the equations.}
\begin{equation}
	\feat_i(x,t)=\frac{1}{n}+
	\sum_{j=0}^{m-1}\int_X dy \ \varphi_{ij}(x,y,t)C_j(y,t)=
	\frac{1}{n}+ (\varphi_t\times C_t)_i(x).
\label{Kernel-contex-def}
\end{equation}
Here we assume that $n$ symbols are generated from the $m$
components of the video. 
Notice that the kernel $\varphi(x,y,t)$ is responsible of expressing the spatial dependencies,
and that one could also extend the context in the temporal dimension. 
However, the immersion in the temporal dimension that arises from the
formulation given in this paper makes it reasonable to begin restricting the 
contextual information to spatial dependencies on the the retina.

\marginnote{Convolutions for built-in incorporation of space-invariance.}
In addition, it is worth mentioning that the agent is expected to return a decision 
also in case of fixed images, which represents a further element for considering 
features defined by Eq.~(\ref{Kernel-contex-def}). 
The filters $\varphi$ can be
regarded as maps from $X\times X\times [0\dts T]$ to $\R^{n,m}$, where
$n$ is the number of the features defined by $\feat$. 
It is worth mentioning that 
whenever $\varphi(x,y,t) \leadsto \varphi(x-y,t)$ the above definition
reduces to an ordinary spatial convolution. The computation 
of   $\feat_i(x,t)$ yields a field with $n$ features, instead of
the three components of color in the video signal.  However, 
Eq.~(\ref{Kernel-contex-def}) can be used for carrying out a piping scheme
where a new set of features $\Feat2$ is computed from $\Feat1$. Of course,
this process can be continued according to a deep computational structure 
with a homogeneous convolutional-based computation, 
which yields the features $C\mskip -10mu\lower-3pt\hbox{\fiverm p}\,$
at the $p$ convolutional layer. 
The theory proposed in this paper focuses on the construction of any of these
convolutional layers which are expected to provide higher and higher abstraction
as we increase the number of layers. 
The {\it filters\/}  $\varphi$ are what completely determines the
features $\feat_i(x,t)$. In this paper we formulate a theory for the discovery of $\varphi$
that is based on three driving principles:

\begin{itemize}
\item	{\em Optimization of information-based indices}\\
	We use an information-based approach  to determine
	$\varphi$. Beginning from the color field $C$, we attach
	symbol $y_i \in \Sigma$ of a discrete vocabulary
	to pixel $(x,t)$ with probability $\feat_i(x,t)$. 
\marginnote{MMI and MaxEnt.}
	The principle of Maximum Mutual Information (MMI) is a natural way 
	of maximizing the transfer of information from the visual source, expressed
	in terms of mixtures of colors, to the source of  symbols $y_i \in \Sigma$.
	Clearly, the same idea can be extended to any layer in the hierarchy.
	Once we are given a certain visual environment over a certain 
	time horizon $[0,T]$\dash---which can  be extended to $[0,+\infty)$\dash---once the filters $\varphi$ have been defined,
	the mutual information turns out to be a functional of $\varphi$, that is denoted
	as $I(\varphi)$. However, in the following, it will be shown that the more general view behind the
	the maximum entropy principle (MaxEnt) offers a better framework for the 
	formulation of the theory.
\item {\em Motion invariance}\\
	While information-based indices optimize the 
	information transfer from the input source $C$ to the symbols,
	the major cognitive issues of invariances are not covered. 
	The same object, which is presented at different scales and under 
	different rotations does require different representations, which transfers 
	all the difficulty of learning to see to the subsequent problems interwound with
	language interpretation. 
	Hence, it turns out that the most important requirement that 
	the visual field $\feat$ must fulfill is that of exhibiting the typical 
	cognitive invariances that humans 
	and animals experiment in their visual environment. 
	We claim that there is only one such fundamental invariance, 
	namely that of producing the same representation for moving pixels. 
	\marginnote{Classic invariances as motion invariance.}
	This incorporates 
	classic scale and rotation invariances in a natural way, which is what is experimented 
	in newborns. Objects comes at different scale and with different rotations simply 
	because children experiment their movement and manipulation. As we track moving pixels, 
	we enforce consistent labeling, which is clearly far more general than 
	enforcing scale and rotation  invariance. 
	The enforcement of motion constraint is the key for the construction of a 
	truly natural invariance. It will be pointed out that motion invariance can always be expressed
	as the minimization of a functional $M(\varphi)$.
\item {\em Parsimony principle}\\
	Like any principled formulation of learning, 
	we require the filters to obey the parsimony principle. 
	Amongst the philosophical implications, it also favors the development of a unique solution.
	The development of filters that are consistent with the above principles requires 
	the construction of an on-line learning scheme, 
	where the role of time becomes of primary importance. 
	The main reason for such a formulation is the need of imposing the development of 
	motion invariance features. Given the filters $\varphi$, 
	there are two parsimony terms, one 
	$P(\varphi)$, that penalizes abrupt spatial
	changes, and another one, $K(\varphi)$  that penalizes 
	quick temporal transitions.
\end{itemize}
%
%
\begin{figure}
		\centering
		\includegraphics{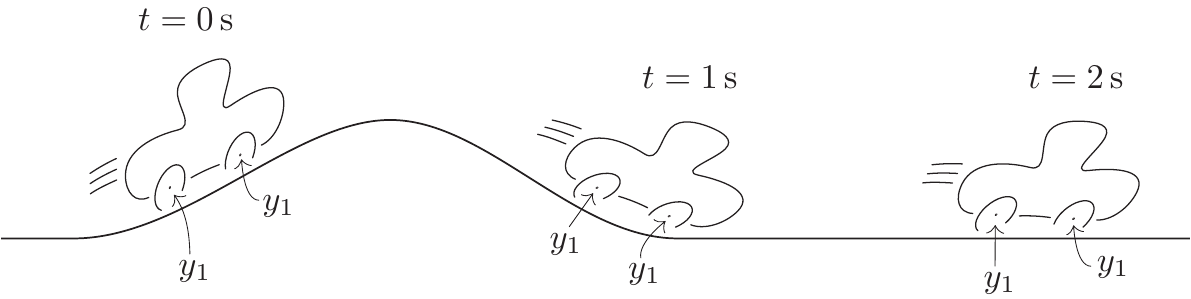}
		\caption{Motion invariance in the feature extraction process.
		The symbol $y_{1}$, that defines a features at the beginning of
		motion ($t=0$ s), must be coherently extracted during the 
		movement\dash---see the enforcement of the coherence requirement at $t=1,2$ s.}
	\label{MotionInvarianceFig}
\end{figure}
\marginnote{Minimization of the cognitive action.}
Overall, the process of learning is regarded as the minimization of the {\em cognitive action}
\begin{equation}
	{\cal A}(\varphi)= - {\cal I}(\varphi) + \lambda_{M} {\cal M}(\varphi)
	+\lambda_{P} {\cal P}(\varphi)
	+ \lambda_{K} {\cal K}(\varphi),
\label{CognitiveActionEq}
\end{equation}
where $\lambda_{M},\lambda_{P},\lambda_{K}$ are positive multipliers.
While the first and third principles are typically adopted in classic unsupervised learning, 
motion invariance does characterize the approach followed in this paper.
Of course, there are visual features that do not obey the motion invariance principle. 
Animals easily estimate the distance to the objects in the environment, a property that clearly
indicates the need for features whose value do depend on motion. The perception of
vertical visual cues, as well as a reasonable estimate of the angle with respect to the
vertical line also suggests the need for features that are motion dependent. 
Since the above action functional ${\cal A}(\varphi)$ depends on the choice
of the multipliers $ \lambda_{M},  \lambda_{P},  \lambda_{K}$, it is quite clear that
there is a wide range of different behavior that depend on the relative weight
that is given to the terms that compose the action. As it will be shown in the following, 
the minimization of ${\cal A}(\varphi)$ can be given an efficient computational
scheme only if we give up to optimize the information transfer in one single step
and rely on a piping scheme that clearly reminds deep network architectures. 
While this paper focuses on unsupervised learning, it is worth mentioning that
the purpose of the agent can naturally be incorporated into the minimization of 
the cognitive action given by Eq~(\ref{CognitiveActionEq}). 

Now, we provide arguments to support the principled framework of this paper. 
Like for human interaction, visual concepts are expected to be acquired by the agents solely by processing their own visual stream along with human supervisions on selected pixels, instead of relying on huge labelled databases. In this new learning environment based on a video stream, any intelligent agent willing to attach semantic labels to a moving pixel is expected to take coherent decisions with respect to its motion. Basically, any label attached to a moving pixel has to be the same during its motion. Hence, video streams provide a huge amount of information just coming from imposing coherent labeling, which is likely to be the primary information associated with visual perception experienced by any animal. Roughly speaking, once a pixel has been labeled, the constraint of coherent labeling virtually offers tons of other supervisions, that are essentially ignored in most machine learning approaches working on big databases of labeled images.
It turns out that most of the visual information to perform semantic labeling comes from the motion coherence constraint, which explains the reason why children learn to recognize objects from a few supervised examples. The linguistic process of attaching symbols to objects takes place at a later stage of children development, when he has already developed strong pattern regularities. We conjecture that, regardless of biology, the enforcement of motion coherence constraint is a high level computational principle that plays a fundamental role for discovering pattern regularities. 
%
%
Concerning the MMI principle, it is worth mentioning that it can be regarded as a 
special case of the MaxEnt principle when the constraints correspond with the
soft-enforcement of the conditional entropy, where the weight of its associated penalty
is the same as that of the entropy (see e.g.~\cite{DBLP:journals/tnn/MelacciG12}). 
Notice that while the maximization of the mutual information nicely addresses the
need of maximizing the information transfer from the source to the selected alphabet of
symbols, it does not guarantee temporal consistency of this attachment. Basically, 
the optimization of the index is also guaranteed by using the same symbol for different
visual cues. Motion consistency faces this issue for any pixel, even if it is fixed.
As for the adoption of the parsimony principle
in visual environments, we can use appropriate 
functionals to enforce both the spatial and temporal smoothness of the solution. 
While the spatial smoothness can be gained by penalizing solutions with 
high spatial derivatives\dash---including the zero-order
derivatives\dash---temporal smoothness arises from the introduction of
kinetic energy terms which penalizes high velocity and, more generally, 
high temporal derivatives.

Since the optimization is generally formulated over arbitrarily large time horizons, 
all terms are properly weighted by a {\em discount factor} that leads to ``forget''
very old information in the agent life. This contributes to a well-position of the
optimization problem and gives rise to dissipation processes~\cite{DBLP:journals/tcs/BettiG16}.






The agent behavior turns out to be driven by the minimization of an 
appropriate functional that combines the all above principles.
The main result in this paper is that this optimization can be interpreted in terms 
of laws of nature expressed by a temporal differential equation. 
When regarding the retina as a discrete structure, we can compute the
probability that at time $t$, in pixel $x$, the emitted symbol is $y_{i}$
by  $1/n +\varphi_{iky}(t) C_{k(x-y)}(t)$. Here, for any pair of symbols
$y_k$, $y_i$, and for any pixel with position $z$, in the coordinate system defined by $x$,
the filter $\varphi_{ikz}$ is the temporal function that the agent is expected
to learn from the visual environment. Basically, the process of learning consists
of determining 
\[
\hat{\varphi} = \arg \min_{\varphi} {\cal A}(\varphi).
\]
In Section~\ref{Cognitive-action-section} we prove that there is no local solution
to this problem, since any stationary point of this functional turns out to be characterized
by the integro-differential equation~(\ref{IntegroDiffEL}). We also show that we can naturally gain 
a local solution when introducing focus of attention mechanisms. Its  purpose is 
to provide a weighed contribution of the single terms of the action by attaching higher 
weights to pixels where the agent is focussing attention. 
Under this re-stating of the problem, we prove that the minimum of the cognitive action 
corresponds with the discovery of the filters $\varphi_{ijy}$
that satisfy the time-variant differential equation
\marginnote{Fourth-order Euler-Lagrange model of learning.}
\begin{equation}
	q^{(4)}(t)+A(t)q^{(3)}(t)+B(t) q^{(2)}(t)+C(t) q^{(1)}(t)+D(t)q(t)+F(t,q(t))=0,
\label{ForthOrderDiff}
\end{equation}
where $q$ is the linearized vector of $\varphi_{ijx}$, matrices $A$, $B$, $C$, $D$, $F$
depend on the time through the video signal and  the trajectory of the focus of
attention, and $F$ is a bounded nonlinear vector.
The subsequent analysis will provide clear evidence on the need of a fourth-order
differential equation for the determination of the filters.
Equation~(\ref{ForthOrderDiff}) has quite a complex structure, since it also contains
the non-linear term $F(t,q(t))$ that, however, as it will be shown is 
piece-wise linear. It is shown that the dependence on time  of the coefficients is inherited by the time-variance of the video. Hence, the solution of the differential 
equation involves dynamics whose spectrum is induced by the video. 
The analysis carried out in the paper shows how can we attack the problem
either in the case in which the agent is expected to learn from a given video
stream with the purpose to work on subsequent text collections, or in the
case in which the agent lives in a certain visual environment, where there is
no distinction between learning and test phases. Basically, it is pointed out
that only the second case leads to a truly interesting and novel result.

\marginnote{Learning by blurring trajectories.}
In particular, the solution of the above differential equation is strongly
facilitated when performing an initial blurring of the video that lasts
until all the visual statistical cues are likely been presented to the agent. 
This very much resembles 
early stages of developments in newborns~\cite{BraddickVR-2011}.  
In so doing, at the beginning,  the coefficients of Eq.~(\ref{ForthOrderDiff})
are nearly constant. In this case, the analysis of the equations leads to
conclude that only a very slow dynamics takes place, which means that
all the derivatives of $q$ are nearly null and, consequently, $q$ is 
nearly constant. This strongly facilitates the numerical solutions and,
in general, the computational model turns out to be very robust, a property
that is clearly welcome also in nature. As time goes by, while the 
blurring process increases the visual acuity the coefficients of the
differential equation begin to change with velocity that is connected
with motion. However, in the meantime, the values of the filters 
have reached a nearly-constant value. Basically, the learning trajectories
are characterized by the mentioned nearly-null derivatives, a condition
that, again strongly facilitates the well-position of the problem. 

%
%
%
A further intuitive reason for a slow dynamics of $q(t)$ is also a consequence of visual invariant features.
For example, when considering a moving car and another one of the same type parked somewhere in
the same frame, during the motion interval, 
the processing over the parked car would benefit from a nearly constant 
solution.  This suggests also searching for the 
same constant solution on the corresponding moving pixel. 
When regarding the problem of learning in a truly on-line mode,
the previous differential equation can be considered as the model
for computing $\varphi_{ijy}$ given the Cauchy conditions.
Of course, the solution is affected by these initial conditions. Moreover,
as it will be clear in the reminder of the paper, the previous differential equations
yield the minimization of the action under appropriate border conditions that 
correspond with forcing a trajectory that satisfies the condition of 
nearly-null of the first, second, and third derivatives of $q$. When 
joined with the blurring process this leads to a causal dynamics
driven by initial conditions that are compatible with boundary conditions
imposed at any time of the agent's life.

The puzzle of extracting robust  cues from visual scenes
has only been partially faced by nowadays successful approaches to computer 
vision. The remarkable achievements  of the last few years have been
mostly based on the accumulation of huge visual collections gathered by
crowdsourcing. An appropriate set up of convolutional networks trained
in the framework of deep learning has given rise to very effective 
internal representations of visual features. They have been successfully used
by facing a number of relevant classification problems by transfer learning. 
Clearly, this approach has been stressing the power of deep learning when 
combining huge supervised collections with massive parallel computation.
In this paper, we argue that while stressing this issue we 
have been facing artificial problems that, from a pure computational point of view, 
are likely to be significantly more complex than natural visual tasks 
that are daily faced by animals. In humans, the emergence of cognition from visual
environments is interwound with language. This often leads to attack
the interplay between visual and linguistic skills by simple models
that, like for supervised learning, strongly rely on linguistic attachment.
However, when observing the spectacular skills of the eagle that catches the pray,
one promptly realizes that for an in-depth understanding
of vision, that likely yields also an impact in computer implementation,
one should begin with a neat separation with language! 
This paper is mostly motivated by the curiosity of addressing a number
of questions that arise when looking at natural visual processes.
While they come from natural observation, they are mostly regarded
as general issues strongly rooted in information-based principles, that we
conjecture are of primary importance also in computer vision.

The theory proposed in this paper offers a computational perspective of
vision regardless of the ``body'' which sustains the processing. 
In particular, the theory addresses some fundamental questions, reported below,
that involve vision processes taking place in both animals and machines. 

\begin{enumerate}

\item [$Q1$] {\em How can animals conquer visual skills without requiring ``intensive supervision''?} \\
Recent remarkable achievements in  computer vision are mostly based
on tons of supervised examples\dash--- of the order of millions! 
This does not explain how can animals conquer visual skills with  scarse
``supervision'' from the environment. 
\marginnote{The call for theories of unsupervised learning.}
Hence, there is plenty of evidence and motivations 
for invoking a theory of truly unsupervised learning 
capable of explaining the process of extraction of features from 
visual data collections. While the need for theories of unsupervised learning 
in computer vision has been advocated in a number 
of papers (see e.g.~\cite{DBLP:journals/corr/TavanaeiMM16},
\cite{Lee:2009},\cite{DBLP:conf/cvpr/RanzatoHBL07},
\cite{DBLP:conf/iccv/GoroshinBTEL15}),
so far, the powerful representations that arise from supervised learning, because
of many recent successful applications, seem to attract much more interest. 
While information-based principles could themselves suffice to construct 
visual features, the absence of any feedback from the environment make those
methods quite limited with respect to supervised learning. Interestingly, the claim
of this paper is that motion
invariance offers a huge amount of free supervisions from the visual environment,
thus explaining the reason why humans do not need the massive supervision
process that is dominating feature extraction in convolutional neural networks.

\item [$Q2$] {\em How can animals gradually conquer visual skills in 
a truly temporal-based visual environment?} \\
Animals, including primates, not only receive a scarse supervision, but
they also conquer visual skills by living in their own visual environment. 
This is gradually achieved without needing to separate learning from test environments. 
At any stage of their evolution, it looks like they acquire the skills that are
required to face the current tasks. On the opposite, most approaches to computer
vision do not really grasp the notion of time. The typical ideas behind on-line learning
do not necessarily capture the natural temporal structure of the visual tasks. Time plays a crucial role in any cognitive process. One might believe that this is restricted to human life, but more careful analyses lead us to conclude that the temporal dimension plays a crucial role in the well-positioning of most challenging cognitive tasks, regardless of whether they are faced by humans or machines. Interestingly, while many people struggle for the acquisition of huge labeled databases, the truly incorporation of
time leads to a paradigm shift in the interpretation of the learning and test environment.
\marginnote{Visual stream can easily surpass any large image collection.}
In a sense, such a distinction ceases to apply, and we can 
regard unrestricted visual collections as the information accumulated during all the agent life, that can likely surpass any attempt to collect image collection. 
The theory proposed in this paper is framed in the context of agent life characterized
by the ordinary notion of time, which emerges in all its facets. We are not concerned
with huge visual data repositories, but merely with the agent life in its own visual environments.

\item [$Q3$] {\em Can animals see in a world of shuffled frames?}\\
One might figure out what human life could have been in a world of visual 
information with shuffled frames. Could  children really acquire visual skills in 
such an artificial world, which is the one we are presenting to machines?
Notice that in a world of shuffled frames, a video requires order
of magnitude more information for its storing than the 
corresponding temporally coherent visual stream. 
This is a serious warning that is typically neglected;
any recognition process is remarkably more difficult when shuffling
frames, which clearly indicates the importance of keeping the spatiotemporal structure
that is offered by nature. This calls for the formulation of a new theory 
of learning capable of capturing spatiotemporal structures. Basically, we need to 
abandon the safe model of restricting computer vision to the processing of images.
The reason for formulating a theory of learning on video instead of on images
is not only rooted in the curiosity of grasping the computational mechanisms 
that take place in nature.
\marginnote{In modern computer vision we have been facing a
problem that is more difficult then that offered by nature.}
 It looks like that, while ignoring the crucial role of 
temporal coherence, 
the formulation of most of nowadays current computer vision tasks leads to
tackle a problem that is 
remarkably more difficult than the one nature has prepared for humans! 
We conjecture that animals could not see in a world of shuffled frames, which 
indicates that such an artificial formulation might led to a very hard problem.
In a sense, the very good results that we already can experiment nowadays are 
quite surprising, but they are mostly due to the stress of the computational power. 
The theory proposed in this paper relies of the choice of capturing
temporal structures in natural visual environments, which is claimed to 
simplify dramatically the problem at hand, and to give rise to lighter computation.

\item [$Q4$] {\em How can humans attach semantic labels at pixel level?}\\
Humans provide scene interpretation thanks to linguistic descriptions. 
This requires a deep integration of visual and linguistic skills, that are
required to come up with compact, yet effective visual descriptions. 
However, amongst these high level visual skills, it is worth mentioning
that humans can attach semantic labels to a single pixel in the retina. 
\marginnote{Pixel-based fundamental primitives.}
While this decision process is 
inherently interwound  with a certain degree of ambiguity, it is remarkably
effective. The linguistic attributes that are extracted are related to 
the context of the pixel that is taken into account for label attachment, while
the ambiguity is mostly a linguistic more than a visual issue. 
The theory proposed in this paper addresses directly this visual skill
since the labels are extracted for a given pixel at different levels of abstraction. 
Unlike classic convolutional networks, there is no pooling; the connection 
between the single pixels and their corresponding features is kept also 
when the extracted features involve high degree of abstraction, that is
due to the processing over large contexts. 
The focus on single pixels allows us to go beyond object segmentation 
based sliding windows, which somewhat reverses the pooling process.
Instead of dealing with object proposals~\cite{DBLP:conf/eccv/ZitnickD14},
we focus on the attachment of symbols at single pixels in the retina.
The bottom line is that human-like linguistic descriptions of visual scenes 
is gained on top of pixel-based feature descriptions that, as a byproduct,
must allow us to perform semantic labeling. Interestingly, there is more;
as it will be shown in the following, there are in fact computational issues
that lead us to promote the idea of carrying our the feature extraction 
process while focussing attention on salient pixels.

\item [$Q5$] {\em Why are there two mainstream different systems in the visual cortex (ventral and
dorsal mainstream)?}\\
It has been pointed out that 
the visual cortex of humans and other primates is composed of two main information pathways
that are referred to as the  ventral stream and dorsal stream~\cite{GoodaleMilner92}.
\marginnote{Is motion invariance the fundamental functional property
that differentiate dorsal and ventral streams?}
The traditional distinction distinguishes the ventral ``what'' and the dorsal ``where/how'' 
visual pathways, so as the ventral stream is devoted to perceptual analysis of the visual input, such as object recognition, whereas the dorsal stream is concerned with providing motion ability in the interaction with the environment. The enforcement of motion invariance is clearly conceived for extracting 
features that are useful for object recognition to assolve the ``what'' task. 
Of course, neurons with built-in motion invariance are not adeguate to make spatial estimations. 
Depending on the the value of the $\lambda_{M}$ parameter, 
the theory presented in this paper leads to interpret the computational scheme of ``ventral neurons'', that
are appropriate for recognition\dash--- high value of $\lambda_{M}$\dash--- or
``dorsal neurons'' that are more appropriate for environmental interactions\dash--- $\lambda_{M}=0$.
The model behind the learning of the filters indicates the need to access to velocity estimation,
which is consistent with neuroanatomical evidence.

\item [$Q6$] {\em Why is the ventral mainstream organized according to a hierarchical architecture with receptive fields?} \\
Beginning from early studies by Hubel and Wiesel~\cite{hubel62}, neuroscientists have
gradually gained evidence of that the visual cortex presents a hierarchical 
structure and that the neurons process the visual information on the basis of inputs 
restricted to receptive field. Is there a reason why this solution has been
developed? We can promptly realize that, even though the neurons are restricted
to compute over receptive fields, deep structures easily conquer the
possibility of taking large contexts into account for their decision.  
\marginnote{Is it there a computational framework to motivates
hierarchical architectures?}
Is this biological solution driven by computational laws of 
vision? In this paper we provide evidence of the fact that receptive fields do 
favor the acquisition of motion invariance which, as already stated, is the 
fundamental invariance of vision. Since hierarchical architectures is the natural
solution for developing more abstract representations by using receptive fields, it turns
out that motion invariance is in fact at the basis of the biological structure of the visual 
cortex. The computation at different layers yields features with progressive degree
of abstraction, so as higher computational processes are expected to use all the
information extracted in the layers.

\item [$Q7$] {\em Why do animals focus attention?}\\
The retina of animals with well-developed visual system is organized
in such a way that there are very high resolution receptors in a restricted
area, whereas lower resolution receptors are present in the rest of the retina.
\marginnote{Is focus of attention driven by computational laws?}
Why is this convenient? One can easily argue that any action typically takes place
in a relatively small zone in front of the animals, which suggests that the 
evolution has led to develop high resolution in a limited portion of the retina. 
On the other hand, this leads to the detriment of the peripheral vision, that 
is also very important. In addition, this could apply for the dorsal system whose
neurons are expected to provide information that is useful to support movement
and actions in the visual environment. The ventral mainstream, with neurons 
involved in the ``what'' function does not seem to benefit from foveal eyes.
From the theory proposed in this paper, the need
of foveal retinas is strongly supported for achieving 
efficient computation for the construction of visual features.  
When looking at Eq.~(\ref{ForthOrderDiff}) it becomes also clear that quick eye movements
with respect to the dynamics of change of the weights of the filters dramatically
simplifies the computation. 

\item  [$Q8$]  {\em Why do foveal animals perform eye movements?}\\
Human eyes make jerky saccadic movements during ordinary visual 
acquisition.  One reason for these movements is that the fovea
provides high-resolution in portions of about $1,2$ degrees.
Because of such a small high resolution portions, the overall sensing of a scene
does require intensive movements of the fovea. Hence, the foveal movements
do represent a good alternative to eyes with uniformly high resolution 
retina. On the other hand, the preference 
of the solution of foveal eyes with saccadic movements is arguable,
since while a uniformly high resolution retina is more complex to achieve
than foveal retina, saccadic movements are less important. The information-based
theory presented in this paper makes it possible to conclude that foveal
retina with saccadic movements is in fact a solution that is 
computationally sustainable and very effective.

\item [$Q9$]  {\em Why does it take 8-12 months for newborns to achieve adult visual acuity?}\\ 
There are surprising results that come from developmental psychology 
on what a newborn see. Charles Darwin came up with the following remark: 
\begin{quote}
	It was surprising how slowly he acquired the power of following with his eyes 
	an object if swinging at all rapidly; for he could not do this well when seven 
	and a half months old.
\end{quote} 
\marginnote{Is there any computational basis of video blurring?}
At the end of the seventies, this early remark was given a technically sound
basis~\cite{DobsonVR-1978}.   In the paper, three techniques, \dash--- optokinetic nystagmus (OKN), preferential looking (PL), and the visually evoked potential (VEP) \dash--- were used to 
assess visual acuity in infants between birth and 6 months of age.
More recently, the survey by  Braddick and Atkinson~\cite{BraddickVR-2011} provides 
an in-depth discussion on the state of the art in the field. It is clearly stated that 
for  newborns to gain adult visual acuity, depending on the specific visual test, 
several months are required.
Is the development of adult visual acuity a biological issue or does it come 
from higher level computational laws? 
This paper provides evidence to conclude that the blurring process taking place
in newborns is in fact a natural strategy to optimize the cognitive action defined
by Eq.~\ref{CognitiveActionEq} under causality requirements.
Moreover, the strict limitations both in terms of spatial and temporal resolution of the
video signal, according to the theory, help conquering
visual skills.  

\item [$Q10$] {\em Causality and Non Rapid Eye Movements (NREM) sleep phases}\\  
Computer vision is mostly based on huge training sets of images, whereas
humans use video streams for learning visual skills. Notice that because of 
the alternation of the biological rhythm of sleep, humans somewhat process collections of
visual streams pasted with relaxing segments composed of ``null'' video signal.
This happens mostly during NREM phases of sleep,
in which also eye movements and connection with visual memory 
are nearly absent. Interestingly, the
Rapid Eye Movements (REM) phase is, on the opposite, similar to ordinary 
visual processing, the only difference being that the construction of visual features
during the dream is based on the visual internal memory representations ~\cite{AndrillonNature2014}. 
As a matter of fact, the process of learning the filters experiments an alternation 
of visual information with the reset of the signal. 
\marginnote{Day-night rhythm and relaxation of system dynamics.}
We provide evidence to claim that 
such a relaxation coming from the reset of the signal nicely fits  the 
purpose of optimizing an overall optimization index based on the 
previously stated principles. In particular, we point out that periodic resetting of the visual 
information favors the optimization under causality requirements.
Hence, the theory offers an intriguing interpretation
of the role of eye movement and of sleep for the optimal development of visual features. In a sense, the theory offers a general framework for interpreting
the importance of the day-night rhythm in the development of
visual features. When combined with newborns blurring, it 
contributes to a relaxation dynamical process that turns out to be
of fundamental importance for the final purpose of optimization
of the visual constraints.

\end{enumerate}


\section{Visual constraints}
%
%
\medskip\noindent
We can provide an interpretation of the processing carried out
by our visual agent in the framework of information theory. 
The basic idea is that the agent produces a set of symbols from a given alphabet while
processing the video.

\medskip\noindent
\textbf{MMI principle.\enspace} 
Let us  define random variables $X$ and $T$, which  take into account the spatial and temporal
probability distribution, while $Y$ is used to specify the probability distribution 
over the possible symbols, and $F$ to specify the frames.
In order to assess the information transfer from $X,T,F$ to $Y$ 
we consider the corresponding mutual information $I$. Clearly, it is zero
whenever random variable $Y$ is independent of $X$, $T$ and $F$.
The mutual information can be expressed by
\begin{equation}
	I(Y;X,T,F)=S(Y)-S(Y\mid X,T,F).
\label{MI-defition}
\end{equation}
The conditional entropy $S(Y\mid X,T,F)$ is given by 
\begin{equation}
	S(Y\mid X,T,F)=-\int_\Omega \sum_{i=1}^n dP_{X,T,F} \  p_i
	\log p_i\, 
\label{CondEntropyDef}
\end{equation}
where $p_i$ is the conditional probability of $Y$ conditioned
to the values of $X$, $T$ and $F$, $dP_{X,T,F}$ is the joint measure of
the variable $X,T,F$, and $\Omega$ is a Borel set in the $(X,T,F)$
space. The agent generates symbols  $y_{i}, \ i=1,\ldots,n$ along with the
corresponding probabilities on the basis of input source that is based on $m$
symbols that are still given along with their probability. 
Now, let us make two fundamental assumptions:
\begin{itemize}
\item The conditional probability $p_i(x,t,C)$, where $C$ is a realization of
random variable $F$,  is given by the $i$-th feature field $\feat_i(x,t)$.
\item Random variables $X,T,F$ follows the ergodic-like assumption, 
\marginnote{Ergodic assumption: probabilistic indices emerge while living in ``wild visual environments"}
so as 
we can perform the replacement:
$$\int_\Omega dP_{X,T,F}\longrightarrow\int_Dd\mu.
$$
\end{itemize}
In what follows we will assume that the measure $d\mu(x,t)$ is
$f(x,t)\,dx\,dt$. Moreover, we assume that $f(x,t)$ is factorized
according to 
\begin{equation}
	f(x,t)=g(x-a(t))h(t),
\end{equation} 
where $a(t)$ is the trajectory of the focus of attention and
$h$ is monotonic increasing function. This ergodic translation
of the probabilistic measure suggests that we pay attention
where the eye is focussing attention, that is in the neighborhood
of $a(t)$: This can be achieved by means of a function $g(x-a(t))$
peaked on the focus of attention. 
\marginnote{Ergodic translation: more weight on 
pixels of focus of attention and on ``recent visual cues.''}
Such a trajectory is assumed to be available 
but, as pointed out in Section~\ref{Discussion-section},
it can also be determined in the overall framework of the
theory presented in this paper. In addition, ergodicity 
here means that we pay attention mostly on ``recent
visual life.'' Clearly, this very much depends on the 
choice of $h$. 
It is quite obvious that the measure 
$d\mu = g(x-a(t))h(t) dx dt$ only makes sense provided
that the function $h$ does not change significantly during 
statistically significant portions of visual environments.  
Whenever these two assumptions hold, 
we can rewrite the conditional entropy defined by
Eq.~(\ref{CondEntropyDef}) as 
\begin{equation}
	S(Y\mid X,T,F)=-\int_Dd\mu(x,t) \ \sum_{i=1}^n\feat_i(x,t)\log
	\feat_i(x,t).
\label{CondEntrbyC}
\end{equation}
Similarly for the entropy of the variable $Y$ we can write
\begin{equation}
	S(Y)=-\sum_{i=1}^n \Pr(Y=y_i)\log\Pr(Y=y_i).
\label{BalEntropybyPr}
\end{equation}
Now, if we use the law of total probability to express $\Pr(Y=y_i)$ 
in terms of the conditional probability $p_i$ and  use
the above assumptions we get
\begin{equation}
	\Pr(Y=y_i)=\int_\Omega dP_{X,T,F} \ p_i=\int_D d\mu(x,t) \, \feat_i(x,t).
\label{BiGPfromp}
\end{equation}
Then
\begin{equation}
	S(Y)=-\sum_{i=1}^n \Big(\int_D d\mu(x,t) \, \feat_i(x,t)\Big)
	\log\Big(\int_D d\mu(x,t) \, \feat_i(x,t)\Big).
\label{BalancingEntropybyC}
\end{equation}
Finally the mutual information becomes
\marginnote{Mutual information based on probabilities $\feat_i(x,t)$.}
\begin{equation}
	I(Y;X,T,F)=\sum_{i=1}^n\Big(\int_D d\mu \, \feat_i\log
	\feat_i - \int_D d\mu \,\feat_i \cdot \log\int_D d\mu \,\feat_i\Big).
\label{MIbyC}
\end{equation}
Of course, $\forall x,t: \ \feat_i(x,t)$ is subject to the probabilistic constraints
\begin{align}
\vcenter{\halign{$#$\hfil&\qquad#\hfil\cr
	\sum_i \feat_i(x,t)=1&(normalization)\cr
	0\le\feat_i(x,t)\le1&(positivity)\cr}}
\label{ProbNormalization}
\end{align}

\medskip\noindent
\textbf{MaxEnt principle.\enspace} 
\marginnote{A more general view: visual constraint satisfaction while maximizing the entropy.}
An agent driven by the MMI principle can carry out an unsupervised
learning process aimed at discovering the symbols defined by random
variable $Y$. Interestingly, when the constraints are given a soft-enforcement, 
the MMI principle has a nice connection with the Max-Ent principle~\cite{Jaynes1957}:
The maximization of the mutual information corresponds with the maximization of 
the entropy while softly-enforcing the constraint that the conditional entropy is null. 
While both the entropy terms get the same absolute value of the weight, 
once can think of different implementations of the MaxEnt principle that very much depend
on the special choice of the weights. When shifting towards the MaxEnt principle
one is primarily interested in the satisfaction of the conditional entropy constraint, while
bearing in mind that the maximization of the entropy protects us from the development of
trivial solutions (see~\cite{MachineLearningMG2018} pp. 99--103 for further details).
Of course, the probabilistic normalization  constraints stated by Eq.~\ref{ProbNormalization} comes along
with the conditional entropy constraint.
The computational mechanism that drives the discovery of the symbols described in this paper is based on MaxEnt, but instead of limiting the unsupervised process to the fulfillment of the conditional entropy constraint, 
we enrich the model with other environmental constraints. 

	First, we notice that the map which originates the symbol production mechanism has not ben given any guideline. The conditional entropy constraint only involves the
value taken by $\feat_{i}$ which depends on  $\varphi_{ij}(x,t)$, but there is no 
structural enforcement on the function $\varphi_{ij}$; its spatiotemporal changes are
ignored. 
\marginnote{Spatiotemporal regularization can be interpreted as  
constraints in the framework of MaxEnt.}
Ordinary regularization issues suggest to discover functions $\varphi_{ij}$ such that
\[
  \lambda_P{\cal P}+\lambda_K{\cal K}=\frac{\lambda_{P}}{2}
  \int_D dtdx\, h(t) (P_{x}
  \varphi_{ij}(x,t))^{2}
  +\frac{\lambda_{K}}{2} \int_{D} dtdx\, h(t)
(P_{t} \varphi_{ij}(x,t))^{2},\]
is ``small'', where $P_{x}, P_{t}$ are spatial and temporal differential operators, and  $\lambda_{P},  \lambda_{K}$ are non-negative reals. Notice that the ergodic translation of
$d\mu$, in this case, only involves the temporal factor
$h(t)$.

	Second, as already pointed out, many relevant visual features need to be motion invariant.
Just like an ideal fluid is adiabatic\dash---meaning that the entropy
of any particle fluid remains constant as that the particles move about
in space\dash---in a video, once we have assigned the correct symbol to a
pixel, due to the fact that the movement of object is continuous, that
symbol is conserved as the object moves on the retina. If we focus attention
on a the pixel $x$ at time $t$, which moves according to the trajectory $x(t)$
then  $\feat(x(t),t) = c$, being $c$ a constant. 
This ``adiabatic'' condition is thus expressed by the condition
$d\feat/dt=0$, which yields 
\begin{equation} 
 	\partial_t \feat_i+v_j\partial_j \feat_i=0,
\label{LocalMIEq}
\end{equation}
where $v\colon D\to\R^2$ is the velocity field that we assume that is given,
and $\partial_k$ is the partial derivative with respect to $x_k$.
When replacing $\feat_i$ as stated by Eq.~(\ref{Kernel-contex-def}) we get
\marginnote{Motion invariance is a linear constraint in
the filter functions $\varphi_{ij}$.}
\begin{align*}
	\int_Xdy\,\big(\partial_t\varphi_{ij} C_j+\varphi_{ij}
	\partial_t C_j +  v_k \partial_k \varphi_{ij} C_j\big)=0,
\end{align*}
which holds for any $(t,x) \in D$.
Notice that this constraint is linear in the field $\varphi$.
This can be interpreted by stating that learning under
motion invariance consists of determining elements of the
kernel of the function 
$\mathscr{M}(\varphi_{ij}):= \int_Xdy\,\big(\partial_t\varphi_{ij} C_j+\varphi_{ij} \partial_t C_j +  v_k \partial_k \varphi_{ij} C_j\big)$. A discussion on the problem of determining 
the kernel of $\mathscr{M}$ is given in~\cite{MachineLearningMG2018}.



\section{Cognitive action}
\label{Cognitive-action-section}
\marginnote{MaxEnt as the minimum of the cognitive action.}
In the previous section we have proposed a method to determine the filters
$\varphi_{ij}$ based on the MaxEnt principle. We provide a soft-interpretation
of the constraints, so as the adoption of the principle corresponds with the minimization
of the  ``action''
\begin{equation}
\begin{split}
{\cal A}_0(\varphi)=& \int_D d\mu \,\feat_i(\varphi) \cdot\log\int_D
d\mu \, \feat_i(\varphi)- \lambda_{C} \int_D d\mu
\,\feat_i(\varphi)\log \feat_i(\varphi)\\ &+\lambda_{1} \int_D \,d\mu
\, \Bigl(\sum_{i=1}^n\feat_i(\varphi)-1\Bigr)^2 -\lambda_{0} \int_D
d\mu \, \feat_i(\varphi)\cdot[\feat_i(\varphi)<0] \cr &+ \frac{\lambda_{P}}{2}
\int_D dtdx\, h(t) (P_{x} \varphi_{ij}(x,t))^{2} +
\frac{\lambda_{K}}{2} \int_D dtdx\, h(t) (P_{t}
\varphi_{ij}(x,t))^{2} \\ &+\lambda_{M} \int_D d\mu \,
\bigl(\partial_t \feat_i(\varphi)+v_j\partial_j \feat_i(\varphi)
\bigr)^2,\\
\end{split}
\label{IntegroDiffEL}
\end{equation}
where the notation $\feat_i(\varphi)$ is used to stress the fact that
$\feat_i$ depends functionally on the filters $\varphi$.  Here the
first line is the negative of the mutual information and the constants
$\lambda_{C}, \lambda_{1}, \lambda_{0}, \lambda_{P},\lambda_{K}$, and
$\lambda_{M}$ are positive multipliers.  In the above formula, and in
what follows we will use consistently Einstein summation convention.

We notice that the mutual information (the first line)
is rather involved, and it becomes too cumbersome to be used with a principle of least action.
However, if we give up to attach the information-based terms the
interpretation in terms of bits, we can rewrite the entropies that define the
mutual information as
\[S(Y\mid X,T,F)\to -\int_D d\mu \, \feat_i^2 \quad \hbox{and}\quad S(Y)\to
-\Bigl(\int_D  d\mu \, \feat_i\Bigr)^2.\]
Interestingly, this replacement does retain all the basic properties on the stationary points
of the mutual information and, at the same time,  it simplifies dramatically the
overall action, which  becomes
\begin{align}
\begin{split}
{\cal A}(\varphi)=&
{1\over 2}\Bigl(\int_D d\mu \, \feat_i(\varphi) \Bigr)^2
- {\lambda_{C} \over 2}\int_D  d\mu \, \feat_i^2(\varphi) \cr
&+{\lambda_{1} \over2}\int_D d\mu \, \Bigl(\sum_{i=0}^{n-1}\feat_i(\varphi)-1\Bigr)^2-\lambda_{0}
\int_D d\mu \, \feat_i(\varphi)\cdot[\feat_i(\varphi)<0]\cr
&+ \frac{\lambda_{P}}{2} \int_D dtdx\, h(t)  (P_{x} \varphi_{ij})^{2}
+\frac{\lambda_{K}}{2} \int_D dtdx\, h(t) (P_{t} \varphi_{ij})^{2} \cr
&+{\lambda_{M}\over2} \int_D d\mu \, \bigl(\partial_t \feat_i(\varphi)+v_j\partial_j \feat_i(\varphi)
\bigr)^2,\cr\end{split}
\label{CognitiveActionEq}
\end{align}
We shall\dash---form now on\dash---assume that the fields $\feat_i$
are extracted by convolution, so that $\feat_i(x,t)=
1/n+\int \varphi_{ij}(x-y,t)C_j(y,t)\,dy$. In order to be sure to preserve 
the commutativity of convolution\dash---a property that in general holds when 
the integrals are extended to the entire plane\dash---we have to make assumptions 
on the retina and on
the domain on which the filters are defined.
First of all assume that $X=[0\dts L]\times[0\dts L]$ and define
$X_4:=[-L\dts L]\times [-L\dts L]$. If the video has support on $X$,
then it is convenient to assume that $\varphi$ has spatial support on
$X_4$; in doing so the commutative property of the convolution is maintained
if we perform the integration on $X_4$. Hence, we have
\begin{equation}
\feat_i(x,t)=\frac{1}{n}+\int_{X_4}\varphi_{ij}(x-y,t) C_j(y,t)\, dy
=\frac{1}{n}+\int_{X_4}\varphi_{ij}(y,t) C_j(x-y,t)\, dy.\label{conv-comm}
\end{equation}
\marginnote{Expression of the variation of the action.}
In what follows we assume that $D\gets X_4\times[0\dts T]$.

The Euler-Lagrange equation of the action arises from $\delta{\cal A}(\varphi)/
\delta \varphi_{ij}(x,t)=0$. So we need to take the variational derivative
of all the terms of  action in Eq.~(\ref{CognitiveActionEq}).
In the following calculation, we will assume  that $d\mu(x,t)=f(x,t)\, dx\,dt$.
The first term yields
\begin{equation}\begin{split}
\int_D \feat_k\, d\mu&\cdot {\delta\over\delta
\varphi_{ij}(x,t)}\int dz\,d\tau\,dy\,f(z,\tau)\varphi_{kj}(y,\tau)
C_j(z-y,\tau)\cr
&=\int dz\, f(z,t) C_j(z-x,t)\cdot\int dz\,d\tau\, d\xi\, f(z,\tau)
\varphi_{ik}(\xi,\tau)C_k(z-\xi,\tau);\cr\end{split}
\label{first-term-variation-non-local}
\end{equation}
while the second term gives
\begin{equation}\begin{split}
&{\delta\over\delta \varphi_{ij}(x,t)}{1\over 2}\int_D \feat_k^2(z,\tau)
 f(z,\tau)\, dz\,d\tau=
\int_X dz\, f(z,t)\feat_i(z,t)C_j(z-x,t)\cr
=&\int d\xi\,\Big(\int dz\,f(z,t)C_j(z-x,t)C_k(z-\xi,t) \Big)
\varphi_{ik}(\xi,t).\cr\end{split}\end{equation}
The variation of the third term similarly yields
\begin{equation}\begin{split}
    \sum_{m=0}^{n-1}\int d\xi\,\Big(\int dz\, f(z,t) C_j(z-x,t) C_k(z-\xi,t)\Big)
    &\varphi_{mk}(\xi,t)\\
    &-\int dz\, f(z,t) C_j(z-x,t).\\
  \end{split}\end{equation}
The variation of the terms that implements positivity is a bit more tricky:
\[\mskip -6 mu\begin{split}
    {\delta\over\delta \varphi_{ij}(x,t)}\int_D \feat_k\cdot[\feat_k<0]\,d\mu&=\int
    {\delta\feat_k(z,\tau) \over\delta \varphi_{ij}(x,t)}\cdot[\feat_k(z,\tau)<0]f(z,\tau)\,dz\,d\tau\cr
    &+\int \feat_k(z,\tau)\cdot{\delta[\feat_k(z,\tau)<0]\over
      \delta \varphi_{ij}(x,t)}f(z,\tau)\,dz\,d\tau. \cr\end{split}\]
However, the second term is zero since
\[\begin{split}
&\int \feat_k(z,\tau)\delta[\feat_k(z,\tau)<0]f(z,\tau)\,dz\,d\tau=\int dz\,d\tau\,d\xi\varphi_{km}(\xi,\tau)C(z-\xi,\tau)\cr
&\cdot\Bigl([\int d\xi\, \varphi_{km}(\xi,\tau)C_m(z-\xi,\tau)+\epsilon
\int d\xi\, \delta\varphi_{km}(\xi,\tau)C_m(z-\xi,\tau)<0]\cr
&\qquad-[\int d\xi\, \varphi_{km}(\xi,\tau)C_m(z-\xi,\tau)<0]\Bigr).\cr\end{split}\]
The difference of the two Iverson's brakets is always zero unless the
epsilon-term makes the argument of the first braket have an opposite sign with
respect to the second. Since $\epsilon$ is arbitrary small, this
can only happen if $\int d\xi\, \varphi_{km}(\xi,\tau)C_m(z-\xi,\tau)=0$.
Thus in either cases the whole term vanishes. Hence, we get
\begin{align}
	{\delta\over\delta \varphi_{ij}(x,t)}\int_D \feat_k \cdot[\feat_k<0]\,d\mu
=\int dz\, f(z,t)C_j(z-x,t)[\feat_i(z,t)<0].
\label{ProbConsELT}
\end{align}
Finally, the variation of the last term is a bit more involved and yields 
(see Appendix~\ref{VariationMotionTerm}):
\begin{align}
  \int d\xi\bigl(\Xi_{jk}(x,\xi,t)\del_t^2+\Pi_{jk}(x,\xi,t)\del_t+
  \Upsilon_{jk}(x,\xi,t)
	\bigl)\varphi_{ik}(\xi,t),
\label{MotionELT}
\end{align}
where
\begin{align}
\begin{split}
\Xi_{jk}(x,\xi,t)&=-\int_D dz\, f(z,t) C_j(z-x,t)C_k(z-\xi,t); \cr
\Pi_{jk}(x,\xi,t)&=\int_D dz\, \bigl(f(z,t)D_tC_j(z-x,t) C_k(z-\xi,t)\cr
&\qquad\qquad\qquad-\del_t(f(z,t)C_j(z-x,t) C_k(z-\xi,t))\cr
&\qquad\qquad\qquad\qquad +f(z,t)C_j(z-x,t) C_k(z-\xi,t)
\bigr);\cr
\Upsilon_{jk}(x,\xi,t)&=\int_D dz\, \bigl(f(z,t)D_tC_j(z-x,t)D_tC_k(z-\xi,t)\cr
&\qquad\qquad\qquad-\del_t(f(z,t)C_j(z-x,t) C_k(z-\xi,t))\bigr).\cr\end{split}
\label{MI-EL-terms}
\end{align}
In doing all this calculations we have used the commutative property of the
convolution as stated in Eq.~(\ref{conv-comm}), if we had not done this
we would have obtained, in some cases, expressions with an higher degree
of space non-locality (i.e. with more than one integral over $X_4$).
\marginnote{Euler-lagrange integro-differential equations for the cognitive action:
they are  neither local in time, nor in space!}
Then the Euler-Lagrange equations reads:
\begin{equation}\begin{split}
    &\lambda_K
    P_t^\star(h(t)P_t\varphi_{ij}(x,t))+\lambda_P h(t)P_x^\star P_x\varphi_{ij}(x,t)\\
    &+c_j(x,t)\cdot\Bigl(\int d\tau\, d\xi\, c_k(\xi,\tau)
\varphi_{ik}(\xi,\tau)-\nu\Bigr)-\beta\int d\xi\, c^2_{jk}(x,\xi,t)\varphi_{ik}(\xi,t)\cr
&\qquad +\nu\sum_{m=1}^n\int d\xi\, c^2_{jk}(x,\xi,t)\varphi_{mk}(\xi,t)
-\lambda\int dz\, f(z,t)C_j(z-x,t)[\feat_i(z,t)<0]\cr
&\qquad\qquad+\eta\int d\xi\bigl(\Xi_{jk}(x,\xi,t)\del_t^2
+\Pi_{jk}(x,\xi,t)\del_t+\Upsilon_{jk}(x,\xi,t)
\bigl)\varphi_{ik}(\xi,t)=0,\cr\end{split}\label{EL1}\end{equation}
where $c_j(x,t)=\int dz\, f(z,t)C_j(z-x,t)$ and $c_{jk}^2(x,\xi,t)=\int dz\,
f(z,t)C_j(z-x,t) C_k(z-\xi,t)$.

\medskip
\noindent
\textbf{Temporal locality.\enspace} 
\marginnote{Approximate and adjoint variable-based methods for  removing temporal non-locality.}
From Eq.~(\ref{EL1}) we immediately see that the first term in the second line
of this equation is non local in time; this means that the
equations are non-causal and therefore it is impossible to regard them
as evolution equations for the filters $\varphi$. To overcome this problem
we propose two different approaches:
\begin{itemize}
\item Enforce time locality by computing the entropy on frames rather
than on the entire life of the agent:
$$\biggl(\int_D \feat_i(x,t) f(x,t)\, dx dt\biggr)^2\to
\int_0^T dt\,\biggl(\int_X \feat_i(x,t) f(x,t)\, dx\biggr)^2.$$
\item Define a causal entropy
$$s_i(t)=\int_0^t\, d\tau\int_Xdx\, \feat_i(x,t) f(x,t),$$
and insert in the action the time average of this quantity together with 
the constraint that enforces this definition. In this way the entropy
term in the Lagrangian will be replaced with
$${1\over T}\int_0^Ts_i^2(t)\, dt+\alpha\int_0^Tdt\, \biggl(s_i(t)-
\int_0^t\, d\tau\int_Xdx\, \feat_i(x,\tau) f(x,\tau) \biggr)^2.$$
\item Define the same causal entropy of point 2. above but insert
the derivative of the constraint that enforces this definition. In this way the entropy
term in the Lagrangian will be replaced with
$${1\over T}\int_0^Ts_i^2(t)\, dt+\alpha\int_0^Tdt\, \biggl(\dot s_i(t)-
\int_Xdx\, \feat_i(x,t) f(x,t) \biggr)^2.$$
In this way the E-L equations that we derive are automatically local in time.
\end{itemize}

For the moment we will develop the theory using the first assumption. The
variation of this term gives, as expected the local form of 
Eq.~\ref{first-term-variation-non-local}: $c_j(x,t)\cdot\int d\xi\, c_k(\xi,t)
\varphi_{ik}(\xi,t)$.

\medskip
\noindent
\textbf{Space locality.\enspace} 
\marginnote{Adjoint equations to remove spatial locality and 
focus of attention.}
We showed that the Euler-Lagrange equations for our 
theory are 
\[\begin{split}\lambda_K P_t^\star(h(t)P_t\varphi_{ij}(x,t))&
    +\lambda_P h(t)P_x^\star P_x\varphi_{ij}(x,t)\\
&+\int_X d\xi\, \bigl[ c_j(x,t) c_k(\xi, t)-\beta c_{jk}^2(x,\xi,t)\cr
&+\eta\bigl(\Xi_{jk}(x,\xi,t)\partial_t^2+ \Pi_{jk}(x,\xi,t)\partial_t
+\Upsilon_{jk}(x,\xi,t)\bigr)\bigr]\varphi_{ik}(\xi,t)\\
&+\nu\sum_{m=1}^n\int_X d\xi\, c^2_{jk}(x,\xi,t)\varphi_{mk}(\xi,t)
-\nu c_j(x,t)\\
&-\lambda\int dz\, f(z,t)C_j(z-x,t)[\feat_i(z,t)<0]=0, \\
\end{split}\]
and as we can see the unknown fields appear inside
a space integral. Now we ask ourselves if it is possible to make this
equations local in space so that they can be regarded as differential equation.

We found that it is possible to do this ``localization'' exploiting
a crucial property of human vision: The focus of attention.
Once we choose $g$,
we can choose a differential operator $L$ such that $Lg=\delta$.

Now if we define the adjoint function $\Gamma_{ij}(x,\xi,t)$ so that
\begin{equation}
\begin{split}
L \Gamma_{ij}(x,\xi,t)= \frac{1}{g(\xi-a(t))}\bigl[&c_j(x,t) c_k(\xi, t)-\beta c_{jk}^2(x,\xi,t)
+\eta\bigl(\Xi_{jk}(x,\xi,t)\partial_t^2\\
&+ \Pi_{jk}(x,\xi,t)\partial_t
+\Upsilon_{jk}(x,\xi,t)\bigr)\bigr]\varphi_{ik}(\xi,t),
\end{split}\label{first-loc-eq}\end{equation}
and the function $\Lambda_j(x,\xi,t)$ as a solution of 
\begin{equation}L\Lambda_{ij}(x,\xi,t)= \frac{\nu}{g(\xi-a(t))}
  \sum_{m=1}^n c^2_{jk}(x,\xi,t)
\varphi_{mk}(\xi,t),\label{second-loc-eq}\end{equation}
we can rewrite Euler Lagrange equations as
\begin{equation}\begin{split}
    \lambda_K P_t^\star(h(t)P_t\varphi_{ij}(x,t))&+\lambda_P
    h(t)P_x^\star P_x\varphi_{ij}(x,t)
    +\Gamma_{ij}(x,a(t),t)+\Lambda_j(x,a(t),t)
-\nu c_j(x,t)\\
&-\lambda h(t)\int dz\, g(z)C_j(z-x,t)[\feat_i(z,t)<0]=0.\end{split}
\label{E-L-local}\end{equation}
Equations~(\ref{first-loc-eq}), (\ref{second-loc-eq}) and~(\ref{E-L-local}) together
form a system of differential equations.

Notice that the spatial function $g$ that we used here to resolve 
space non-locality is the same function that appears in the measure $d\mu$;
however we could also have chosen a different function.

%
%


\section{Neural interpretation in discrete retina}
\label{discretization-section}
\marginnote{Allocating one neuron per pixel:
The filters $\varphi_{ijx}(t)$ are defined on the quantized retina $X^\sharp$.}
Up to this point we have proposed our theory as a field theory, we now
consider the corresponding theory defined on a discretized retina
$X^\sharp$. For each point $x$ of the discretized retina we then have a
variable $\varphi_{ijx}(t)$. 

Since all the terms in the cognitive action (except for the kinetic terms)
are expressed in terms of the feature field $\feat_i(x,t)$, so we need to show
how this fields can be written on a discretized retina
$X^\sharp=\{(i,j)\mid 0\le i<\ell, \quad 0\le j<\ell\}$. On a discrete retina we
will have instead of the fields $\varphi_{ij}(x,t)$ a bunch of functions of the
time variable $\varphi_{ij x}(t)$, indexed by the point on the retina $x$ other than the
filter indices $i$ and $j$. Similarly the color field will
be replaced by $C_{ix}(t)$.

Using Einstein notation  we have that the discretized form
of the feature fields is $\feat_{ix}^\sharp(t)=1/n+
\varphi_{iky}(t)C_{k(x-y)}(t)$, where the sum over $y$
is performed over the discrete retina $X^\sharp$. Then for example
the two pieces of the motion invariance term becomes
\[\begin{split}
    &\del_t\feat_i(x,t)\mathrel{\mathop{\longrightarrow}^{\rm disc\,\,\,}}
    \dot\feat_{ix}^\sharp(t)=\dot \varphi_{iky}(t)C_{k(x-y)}(t)+
    \varphi_{iky}(t)\dot C_{k(x-y)}(t);\cr
    &\del_j\feat_i(x,t)\mathrel{\mathop{\longrightarrow}^{\rm disc\,\,\,}}
    \Delta_j \feat_{ix}^\sharp(t)= \varphi_{iky}(t)\Delta_jC_{k(x-y)}(t).
\end{split}\]
The term of motion invariance becomes a
quadratic form in $\varphi$ and $\dot \varphi$ since
\begin{equation}\label{MI-quadratic-discrete}\begin{split}
&\varphi_{iky} \big[g_x(v\cdot \Delta C_{k(x-y)}+\dot C_{k(x-y)})
(v\cdot \Delta C_{m(x-z)}+\dot C_{m(x-z)})\delta_{ij}\big]\varphi_{jmz}\\
&+2\varphi_{iky}\big[g_x(v\cdot \Delta C_{k(x-y)}+\dot C_{k(x-y)})C_{m(x-z)}
\delta_{ij}\big]\dot\varphi_{jmz}\\
&+\dot\varphi_{iky}\big[g_x C_{k(x-y)}C_{m(x-z)}
\delta_{ij}\big]\dot\varphi_{jmz}.\end{split}\end{equation}
The other relevant terms of the theory (the entropy, the
relative entropy and the probabilistic constrains) are just a function
$V(\varphi(t),t)$.

\marginnote{Factorization of $f(x,t)=h(t)g(x-a(t))$}
Notice also that because of the proposed factorization
of the weight function $f(x,t)=h(t)g(x-a(t))$ the term $g_x$ in the discretized
formulation is also a function of time, and as it has been pointed out
in Section~\ref{driving-principles}. This contributes to the time dependence
that affects the coefficients of the differential equation that governs
the evolution of the filters. However since $g_x$ plays the role of a
probability distribution over the retina it must be that
$\sum_{x\in X^\sharp}g_x=1$ for every $t$.

\marginnote{Tensor linearization.}
Before going on to describe the theory on the
discretized retina we will show how it is possible to linearize the indices of $\varphi$
in order to deal with a vectorial variable rather than a more complex tensorial index structure.
In order to be more precise in the construction we will split the retina index
$x$ of $\varphi$ into its two discrete coordinates $x_1$ and $x_2$ so that the filters fields
will be identified by four rather than three indices. For the same reason when
considered necessary we will also explicitly write down the summations.
As we have argued the first step towards discretization is
\[\begin{split}
\feat_i(x,t)\longrightarrow \feat_{ix}^\sharp(t)&:=\frac{1}{n}+\sum_{k=0}^{m-1}
\sum_{\scriptstyle -\ell\le\xi_1\le\ell\atop\scriptstyle -\ell\le\xi_2\le\ell}
\varphi_{ik(x_1-\xi_1)(x_2-\xi_2)}(t)C_{k\xi_1\xi_2}(t)\\
&=\frac{1}{n}+\sum_{k=0}^{m-1}
\sum_{\scriptstyle -\ell\le\xi_1\le\ell\atop\scriptstyle -\ell\le\xi_2\le\ell}
\varphi_{ik\xi_1\xi_2}(t)C_{k(x_1-\xi_1)(x_2-\xi_2)}(t).
\\\end{split}\]
In Appendix~\ref{linearization-appendix} we show that the feature field
can be rewritten as
\[ \feat_{ix}^\sharp(t)=\frac{1}{n}+\sum_{j\in J_i}
  q_j(t) \Gamma_{T_x(j)}(t),\]
where $q_j$ are the linearized features and $\Gamma_j$ the linearized input.
The map $T_x$ transforms appropriately $j$ into another index depending on
the point in which the convolution is computed (see
Appendix~\ref{linearization-appendix}).

The cognitive action then can be written with appropriate
regularization terms as
\[
 \begin{split}
  {\cal A}^\sharp(q)={\cal A}_0^\sharp(q)&+
  \Bigl(\int_0^Tdt\, h(t)g_x
  \feat_{ix}^\sharp(t)\Bigr)^2
  \!\!+\int_0^T dt\, h(t)\Bigl[\frac{\lambda_1}{2} g_x \Bigl(\sum_{i=0}^{n-1}\feat_{ix}^\sharp(t)-1
   \Bigr)^2\\
   &-\frac{\lambda_C}{2} g_x (\feat_{ix}^\sharp(t))^2+\frac{\lambda_M}{2}g_x\bigl(\dot\feat_{ix}^\sharp(t)+
   v_k\Delta_k\feat_{ix}^\sharp(t)\bigr)^2\\
   &-\lambda_0 g_x \feat_{ix}^\sharp(t)[\feat_{ix}^\sharp(t)<0]\Bigr],\\
  \end{split}
\]
where ${\cal A}_0^\sharp(q)$ is a suitable regularization term that we will discuss in the
following.

Let us now see how  the action  can be rewritten
in terms of the variables $q$. The motion invariance term becomes (see
Eq.~(\ref{MI-quadratic-discrete}))
\[
  \int_0^T dt\, h(t)\sum_{i=1}^n \sum_{j,l\in J_i}
  \left(\frac{1}{2} \dot q_j M_{jl}(t)\dot q_\ell +q_j N_{jl}(t)\dot q_\ell
  +\frac{1}{2} q_j(t)O_{jl}(t)q_l(t)\right),
\]
where the matrices can be expressed as:
\begin{equation}
\begin{gathered}
M_{j\ell}= g_x \Gamma_{T_x(j)}\Gamma_{T_x(l)},\qquad
N_{j\ell}= g_x (\dot \Gamma_{T_x(j)}+v_k\Delta_k\Gamma_{T_x(j)})
\Gamma_{T_x(l)},\\
O_{jl}=g_x (\dot \Gamma_{T_x(j)}+v_k\Delta_k\Gamma_{T_x(j)})
(\dot \Gamma_{T_x(l)}+v_k\Delta_k\Gamma_{T_x(l)}).
\end{gathered}
\label{MI-matrices}
\end{equation}
As it is explained in
Appendix~\ref{linearization-appendix},
by a careful redefinition of these matrices,
we can transform the sum over $J_i$ as a sum over the entire set $J$ (definitions of
$J$ and $J_i$ are also in Appendix~\ref{linearization-appendix} and they are essentially
subsets of $\{0,\dots,nm(3\ell+1)-1\}$). Thus the motion
invariance term is just
\[\int_0^T dt\, h(t)\left(\frac{1}{2} \dot q_j M^\natural_{jl}(t)\dot q_\ell
    +q_j N^\natural_{jl}(t)\dot q_\ell
    +\frac{1}{2} q_j(t)O^\natural_{jl}(t)q_l(t)\right),\]
and from now on the range of the indices of repeated sums is intended to
be $J$.

Because of the way in which the problem is formulated, it seems natural to chose as
a criterion for the choice of the filters the {\em minimization} of the functional
$\cal A^\sharp$.
For this reason the
regularizing part of the functional ${\cal A}_0^\sharp$ must be carefully chosen
so that it does not spoil the coercivity of the functional, but we need also to be sure that
it will give rise to {\em stable} EL equations.

Coercivity and stability cannot be obtained with a regularization term that contains only
first derivatives in time (see~\cite{DBLP:journals/tcs/BettiG16}).
\[\begin{split}
  {\cal A}_0^\sharp(q)&=\int_0^Tdt\, h(t)\left(\frac{\alpha}{2}|\ddot q|^2
    +\frac{\beta}{2}|\dot q|^2 +\gamma \ddot q\cdot \dot q+\frac{k}{2}|q|^2
  \right)\\
  &=\int_0^Tdt\, h(t)\left(\frac{\bar\alpha}{2}|\ddot q|^2
    +\frac{\bar\beta}{2}|\dot q|^2 +\frac{\gamma}{2}(\ddot q+\dot q)^2
    +\frac{k}{2}|q|^2\right).
\end{split}\]
\marginnote{Cognitive action in its linearized form.}
All the other terms can be discretized as well (see Appendix~\ref{linearization-appendix}
for the details) so that the action as a functional of the $q$s reads
\begin{equation}
  \begin{split}
  {\cal A}^\sharp(q)= &\sum_{i=0}^{n-1}\Bigl(\int_0^Tdt\, h
  \sum_{j\in J_i} b_j q_j\Bigr)^2+ \int_0^Tdt\, h \Bigl(\frac{\alpha}{2}|\ddot q|^2+
  \frac{\beta}{2}|\dot q|^2+\gamma\ddot q\cdot \dot q+
  \frac{\lambda_M}{2}\dot q\cdot M^\natural \dot q\\
  &+\lambda_M q\cdot N^\natural \dot q+\frac{1}{2}q\cdot \bigl(k-\lambda_CM^\natural+
  \lambda_1 M+\lambda_M O^\natural\bigr)q\\
  &+\frac{1-\lambda_C}{n}b\cdot q+w_s(t,q)\Bigr),\\
  \label{disc-action-1-form}
\end{split}
\end{equation}
where $\alpha$, $\beta$, $\gamma$, $k$, $\lambda_M$, $\lambda_1$ and $\lambda_C$ are real positive
constants
while 
\begin{equation}\begin{split}
  b_j(t)&:=\sum_{x\in X^\sharp} g_x \Gamma_{T_x(j)}\\
  w_s(t,q)&:=\sum_{i=0}^{n-1}\sum_{x\in X^\sharp} g_x
    \Bigl(\frac{1}{n} +\sum_{j\in J_i}q_j\Gamma_{T_x(j)}\Bigr)
    \Bigl[\sum_{k\in J_i}q_k\Gamma_{T_x(k)}<-\frac{1}{n}\Bigr].\end{split}
 \label{sign-and-b-term}
  \end{equation}
Then if we formulate our minimization problem on the set  $K:=\{\, q\in H^2(0,T;\R^L)\mid q(0)=q^0,
\dot q(0)=q^1\}$ under the assumption that $h$ is limited in the interval $[0\dts T]$, and that
we can choose $k$ big enough so that the quadratic term in $q$ in Eq.~(\ref{disc-action-1-form})
is positive definite, we can prove that the minimum of the functional ${\cal A}^\sharp$ on the set $K$
exists (the apparently dangerous linear term in $q$ with possible negative coefficient can also
be controlled with the regularization term $k/2 |q|^2$).

\marginnote{Time-local form of the cognitive action in its linearized form; space locality is a 
direct consequence of retina quantization.}
If we also make the entropy term local in time we get
\begin{equation}\begin{split}
  {\cal A}^\sharp(q)=  \int_0^Tdt\, h \Bigl(&\frac{\alpha}{2}|\ddot q|^2+
  \frac{\beta}{2}|\dot q|^2+\gamma\ddot q\cdot \dot q+\frac{\lambda_M}{2}\dot q\cdot M^\natural \dot q
  +\lambda_M q\cdot N^\natural \dot q\\
  &+\frac{1}{2}q\cdot \bigl(k+B^\natural-\lambda_CM^\natural+
  \lambda_1 M+\lambda_M O^\natural\bigr)q\\
  &+\frac{1-\lambda_C}{n}b\cdot q+w_s(t,q)\Bigr),\\\end{split}
\label{action-in-the-discrete}
\end{equation}
where $B_{jl}=b_j b_l$.
As it is remarked above the minimization problem takes place in the convex and closed
set $K$ and then in order to evaluate the
first variation of this functional
we need to take as a varying function $v\in V=\{v\in C^\infty(0,T; \R^L)\mid v(0)=\dot v(0)=0\}$.


The differential E-L equation for the whole functional thus reads:
\marginnote{Forth-order Euler-Lagrange differential equation of 
learning; the term $\nabla_q w_s(t,q)$ is piece-wise linear.}
\begin{equation}\begin{split}
\hat \alpha q^{(4)}+2\dot{\hat\alpha}q^{(3)}+(\ddot{\hat\alpha}+\dot{\hat\gamma}-\hat R)
\ddot q &-\left(\dot{\hat R}-\ddot{\hat\gamma}-\lambda_M\hat N^\natural+\lambda_M(\hat N^\natural)'\right)\dot q\cr
&-\left((\dot{\hat N}^\natural)'-\hat P\right)q +\frac{1-\lambda_C}{n}b+\nabla_q w_s(t,q)=0,
\cr\end{split}
\label{Gen-q-DiffEq}
\end{equation}
here $P:=k+B^\natural-\lambda_CM^\natural+
\lambda_1 M+\lambda_M O^\natural$, $R:=\beta+\lambda_M M^\natural$
and we have used the notation $\hat f=h f$ (so that for example
$\ddot{\hat \alpha}=\ddot h \alpha$).

In deriving equations some conditions arises naturally at $t=T$:
\begin{align}
\begin{split}
\label{BoundCondq0}
  &\hat\alpha\ddot q(T)+\hat\gamma\dot q(T)=0\\
  &\hat\alpha q^{(3)}(T)-\dot{\hat\alpha}\ddot q(T)+(\hat\beta+\lambda_M\hat M^\natural -\dot{\hat\gamma})
    \dot q(T)+\lambda_M(\hat N^\sharp)'q(T)=0
\end{split}
\end{align}
This is in fact a system of second order ODE's with non-constant coefficients.

An interesting special case of these equations is that obtained with a null
signal $C\equiv0$.  With this assumption our equations become
\[\hat \alpha q^{(4)}+2\dot{\hat\alpha}q^{(3)}+(\ddot{\hat\alpha}+\dot{\hat\gamma}-\hat\beta)
\ddot q+(\ddot{\hat\gamma}-\dot{\hat\beta})\dot q+\hat kq=0.\] 
Now assume that $h(t)=e^{\theta t}$, with $\theta$ positive, then
\marginnote{System dynamics at night: $C=0$.}
\begin{equation}
	\alpha q^{(4)}+2\theta\alpha q^{(3)}+(\theta^2 \alpha+\theta\gamma-\beta)
  \ddot q+(\theta^2\gamma-\theta\beta)\dot q+kq=0.
\label{NightDynamics}
\end{equation}
In order to see whether this equation can be stable we need to apply the 
Routh-Hurvitz criterion.
For a fourth order ODE $q^{(4)}+aq^{(3)}+b\ddot q+c\dot q+dq=0$ this criterion reduces to check if
$a>0$, $b>0$, $0<c<ab$ and $0<d<(abc -c^2)/a^2$ that in our case means that
\marginnote{Stability conditions yield a relaxation process
that guarantees the vanishing of the any temporal derivative of $q$.}
\begin{equation}
\alpha>0,\quad \beta>0,\quad k>0\quad \gamma>\frac{\beta}{\theta},\quad
0<\alpha<\frac{(\beta-\gamma\theta)  [\beta-\theta(\gamma+2\theta)]}{4 k}.
\label{StabilityConditions}
\end{equation}
So for example if we choose $\alpha=1/2$, $\gamma=2$ and
$\theta=\beta=k=1$ we obtain a stable equation. This being said it is
also crucial to notice that we have control over the important
parameter of the theory $\theta$ as long as you choose the
regularization parameters carefully. Also $k$ can be chosen large
enough to overcome the negative contribution of the conditional
entropy and to make the overall term positive definite.

We are now in condition to address most of the questions
raised in Section~\ref{driving-principles}.

\medskip
\noindent
\textbf{Addressing $Q1,\ldots,Q8$.\enspace} 
The expression of the minimum  of the cognitive action
leads to computational laws on visual feature in convolutional
networks that very well address the questions 
$Q1$, $Q2$, $Q3$, $Q4$.   
The theory shows that any visual agents\dash---including animals\dash---can conquer visual skills without requiring $(Q1)$ ``intensive supervision.'' In particular, the Euler-Lagrange differential equations that dictate the agent life only process visual streams without any supervision, so as they represent a fully-unsupervised method of feature extraction. The agent interaction with the environment can, later on, at different stage of development,  benefit from a number of different forms of supervision that can refine the features developed according to the proposed scheme. Unlike most approaches from machine learning, here the role of time is of crucial importance $(Q2)$. Interestingly, it becomes clear what is the effect of 
shuffling video frames $(Q3)$, which is likely to be also
a ``cul de sac'' for humans!
Finally, the given laws of feature development inherently operate at pixel level $(Q4)$ so as to support a primitive of vision that is clearly available in humans. The discussion on the role of motion invariance also indicates the reason why we need distinct feature models
depending on the visual purpose. In particular, motion invariance
turns out to be useful for recognition tasks, whereas such a 
property must not hold for neurons involved in motion control of
the agent. This addresses question $Q5$ on the reason 
why human visual cortex is in fact composed of two 
different systems, namely the ventral and dorsal mainstream.
The spatial non-locality of the Euler-Lagrange equations
discussed in the previous section~(\ref{EL1})
suggests that efficient solutions cannot be discovered 
without making assumptions on the convolutional filters.
Interestingly, the introduction of adjoint variables makes
is possible to remove spatial locality. This is especially
interesting when the Green function is chosen as the $g$
functions of the probabilistic measure. Clearly, it is a peaked 
bell-shaped structure of $g$, centered on the point
of focus of attention $a(t)$, that leads to discover peaked
filters. This arises from simple arguments: in case of peaked
$g$ the convolutional net only needs to react to pixels
close to the focus of attention and the spatial regularization
does the rest. Hence, neurons that are based on receptive
fields become the natural solution and, consequently, 
we nicely also interpret the reason why the 
ventral mainstream is organized according to a hierarchical architecture $(Q6)$. At the same time, also a natural answer
arises on the reason why do animals with very good visual skills
typically focus attention $(Q7)$. In particular, animals with foveal eyes
they perform complex movements with the main purpose
of locating informative regions of the retina. At the light of
the theory proposed in this paper, the factor $g$ that is
used in the ergodic translation of the probabilistic measure 
$d\mu$ plays a crucial role defining a system dynamics to
address $(Q6)$ and $(Q7)$.


\section{Boundary conditions and causality}


We have seen that the batch-mode approach is motivated by the 
property that, under appropriate conditions, 
the cognitive action is convex and admits a unique minimum.
This holds true when boundary conditions~(\ref{BoundCondq0}) are imposed. 
Interestingly, these conditions are verified in the simple case in which
\begin{align}
\marginnote{Boundary ``relaxing conditions.''}
\label{q-InitialValue}
	q(0)=0 \\ \dot{q}(T) =  
	\ddot{q}(T)=q^{(3)}(T)=0
\label{ZeroBorderCond}
\end{align}
Clearly, the conditions $q(0)=0$ 
and $\dot{q}(0)=0$ are also a natural choice; in particular, from the first
one we conclude that $\forall x \in X: \ \ \feat_{i}(x,0)=1/n$. 
Throughout the paper, these are referred to as the ``relaxing boundary conditions.''
Now, if we want to gain causality we need to impose that the optimal
solution is one which is independent of the time horizon. Suppose we want to
discover a solution $q$ over $[0\dts t_{2}]$ such that, once 
restricted to $[0\dts t_{1}]$ is the same as the solution  over $[0\dts t_{1}]$
for any $t_{2}>t_{1}$. For this to happens, we can promptly see
that Euler-Lagrange equations to be used for solving the problem over $[t_{1}
\dts t_{2}]$
do require to be initialized according to Cauchy conditions on $t_{1}$. This 
arises when considering that as $t_{2} \rightarrow t_{1}$ the above boundary conditions
for the solution  over $[t_{1}\dts t_{2}]$ collapse to Cauchy conditions. 
Clearly, forcing the initial conditions is not necessarily consistent with 
searching the minimum under boundary conditions~(\ref{BoundCondq0}). 
The consistence, however, can be gained whenever we conceive a novel learning
scheme which corresponds with a temporal deformation of the given Lagrangian. 
Let us focus on the special case in which the boundary conditions~(\ref{ZeroBorderCond}) are 
imposed. In order to guarantee the consistency
with Cauchy initial conditions, one needs to ensure the fulfillment of the condition
on the derivatives of $q$ on the right border, which can be if we upper bound
the values of $\Vert q^{(i)}\Vert$. This is made possible by the 
very special nature of the video signal $C$; if we scale it up and down, we do not
change the associated information. Hence, If $0 \leq \rho \leq 1$ then 
$C_{b}=\rho C$ carries out the same information as $C$. 
However, notice that as $\rho \rightarrow 0$,
because of the retina sensibility, the signal is lost. When quantizing  $C_{b}$, 
information loss is due to the limited number of bits used to represent $C$. Now, 
let $\epsilon_{1},\epsilon_{2},\epsilon_{3}>0$ be derivative thresholds to ensure
that we follow a learning trajectory that is {\em nominally compatible} with the
relaxing boundary conditions~(\ref{ZeroBorderCond}). Basically we assume that, instead of feeding the machine
with signal $C$, the input is $C_{d}$ defined as
\begin{align}
\marginnote{Blurring process $\rho$ dynamics and 
day-night rhythm.}
\label{Cbeq}
	C_{b} = \prod_{i=1}^{3}[\epsilon_{i} - \Vert q^{(i)} \Vert >0]  \ \rho C\\
	\dot{\rho} = \eta (1-\rho)_{+} \prod_{i=1}^{3} (\epsilon_{i} -  \Vert q^{(i)} \Vert)_{+},
\label{ThresholdUpd}
\end{align}
where $(\alpha)_{+}:=[\alpha \geq 0] \alpha$ and 
$\rho(0)=0$. We can easily see that the $\rho$ dynamics
is consistent with $0 \leq \rho \leq 1$. Likewise, the 
learning trajectory that is {\em nominally compatible} with the boundary conditions since
we always have 
$\Vert \dot{q} \Vert < \epsilon_{1}, \ \Vert \ddot{q} \Vert  < \epsilon_{2}$ and  
$\Vert q^{(3)} \Vert < \epsilon_{3}$. This holds true because of the ``night relaxation
dynamics" in case in which Eq.~(\ref{NightDynamics}) is stable, that is
guaranteed when choosing a set of parameters that satisfy the
conditions~(\ref{StabilityConditions}).
As any  of the conditions $\epsilon_{i} -  \Vert q^{(i)} \Vert >0$ is violated, the threshold
$\epsilon_{i}$ is reduced to $\zeta \epsilon_{i}$, that is
$\epsilon_{i} \leftarrow   \zeta \epsilon_{i}$, with $0 \leq \zeta \leq 1$,
so as to carry out a sort of hysteresis process. 
We can easily that the time the agent is sleeping at night is strongly
depending on $\zeta$.
Its choice contributes to define the duration of a relaxation behavior until 
until $|q^{(i)}|  > \zeta \epsilon_{i}$ holds true. As the condition is met the threshold $\epsilon_{i}$
is restored in Eq~(\ref{Cbeq}).
The overall dynamics guarantees that 
$\zeta \epsilon_{i} <  \Vert q^{(i)} \Vert  <  \epsilon_{i}$. 
To sum up, when $\epsilon_{i}$ is small enough the learning trajectory
that is driven by the Cauchy initialization evolves is always kept close
to the boundary conditions on the right side of the interval. 
This guarantees that the learning problem formulated as the minimization
of the cognitive action is solved on any interval, regardless of its length. 
In a sense, there is no difference between the learning and test; the 
agent simply lives in its own learning environment by optimizing the
cognitive action, that is in fact expressing the constraints to be 
satisfied. 

\medskip
\noindent
\textbf{Addressing $Q9$: blurring in newborns.\enspace} 
As already noticed, while the transformation $C \rightarrow C_{b}=\rho C$
does not destroy information in the continuum setting of computation, 
as the video is 	quantized, no matter how many bits we choose,
for arbitrarily small values of $\rho$ the signal $C_{b}$ can be significantly
blurred. Interestingly, the above computational scheme, which is dictated 
only by mathematical requirements of satisfying the boundary conditions,  
is what happens in newborns during their evolutive process of 
vision acquisition. It is worth mentioning that the scale transformation
$C \rightarrow C_{b}=\rho C$ is not the only one which supports
a system dynamics that is nominally compatible with the boundary conditions. 
If the signal $C$ undergoes a spatiotemporal low-pass filtering computation
we end up in conclusions that the correspondent dynamics is 
very similar to those driven concerning Eqs.~(\ref{Cbeq}) 
and~(\ref{ThresholdUpd}).
Notice that video blurring, which induces a 
deformation into the Lagrangian
favors the development of a technically sound solution and it is
at the same time fully compatible with the problem of learning in
visual environments. The reason is that humans and machines 
must do their best to extract visual cues also in presence of 
fog or in dark environments. Hence, a learning process focussed 
on information extraction in case of blurred video turns out to be 
useful for the kind of visual skills that are ordinarily required.
\marginnote{Blurring and causal agents.}
A strong property that is acquired by visual agent carrying out 
a learning trajectory that is {\em nominally compatible} 
with the boundary conditions is that they are {\em causal agents}.
This is straightforward consequence of the fact that the 
conditions~(\ref{ZeroBorderCond}) are nominally verified at any
time of the agent life.   The causality of the agent is of crucial 
importance for the learning process, since causality  
has important consequences on statistical consistency. 
The last but not the least, the described blurring process plays
a crucial role to make the Euler-Lagrange differential equations
of learning very-well conditioned. Again, this is in fact a consequence
of imposing a learning trajectory where the derivatives of $q$
are upper bounded with small values. This allows us to set up
a numerical framework with strong precision and, at the same time, 
with limited computational burden. This is also gained because of the 
limitations on $\Vert q^{(i)}\Vert$ and it is clearly an important feature
in computer vision. An accurate analysis on the dynamics of $\rho$
clearly indicates that the choice of $\eta$ plays a crucial role. 
One very much would like to synchronize the end of blurring with 
the end of learning, that must be somewhat connected with the 
statistical structure of the given visual environment. 
It looks like that there is no clear way of perfectly performing such a 
synchronization. The main reason is that we are in front of complex
dynamics driven by the video signal that are likely hard to catch.  
This analysis addresses directly item $Q5$ 
raised in Section~\ref{driving-principles}.


\medskip
\noindent
\textbf{Addressing $Q10$: day-night rhythm.\enspace} 
Interestingly, the described relaxation process is just a way to
give up looking for a magic blurring process and rely, on the 
opposite, on a strategy that guarantees built-in satisfaction of the
boundary conditions. Eqs~(\ref{Cbeq}) and~(\ref{ThresholdUpd})
draws an intriguing picture that very much reminds us of the
day-night human alternation rhythm. Interestingly, the idea of 
following a trajectory that is glued to the boundary conditions 
naturally leads to segment the agent's life into days and nights. 
While the video processing takes place whenever 
$\epsilon_{i} - \Vert q^{(i)}\Vert > 0$, the violation of the condition
corresponds to the agent sleep phase where relaxation takes
place. 
\marginnote{Day-night rhythm is primarily defined
by the choice of $\zeta$. }
Clearly, the day-night alternation rhythm depends strongly on the
choice of $\zeta$. Interestingly, there is no special reason for assuming
that it must be kept constant during the agent's life; other objectives
could drive a different decision. In addition, notice that the agent's
relaxation results in a truly sleep in the sense that the agent,
unless you do not waked it up,
cannot process visual information upon request. 
Interestingly, this gives the agent a truly human-like behavior, that 
pairs with the common framework of learning under video blurring. 

The overall process dynamics is also dependent on other remarkable
choices. In particular, the analysis carried out in this paper focuses
on a special choice of function $h$ (e.g. $e^{\theta t}$). 
Its consequence is that for an effective 
generalization process to take place one needs to establish a 
very slow dissipative processes to guarantee that the weighting term $h$ does not 
penalize too much back in time.  This is of crucial importance in order to
give the agent the capability of storing enough information for its decision.

This analysis addresses directly directly item $Q6$ 
raised in Section~\ref{driving-principles}.




\section{Discussion}
%
\label{Discussion-section}
This paper is about information-based laws of visual features, that are determined under
the principle of Maximum Entropy.
\marginnote{Learning in wild visual environments.}
Unlike most machine 
learning approaches to computer vision, here we assume that the agent which performs
feature extraction lives in an ``wild visual environment,'' and processes 
video streams instead of huge image repositories.  Amongst the constraints
imposed in the convolutional-based feature generation, motion invariance does represent
the major distinctive idea. 

\marginnote{Motion invariance as the truly unique invariance.}
It is pointed out that all other invariances, that are typically 
considered in models of computer vision, turn out to emerge naturally from imposing
feature motion invariance. In a sense, imposing such an invariance 
corresponds with the natural exploitation of a huge amount of supervised
information from nature, which invites us to propose a coherent labelling
for moving pixels of a certain object. 
The proposed theory makes it possible to learn the filters
of convolutional nets by solving differential equations that can be regarded as
information-based laws of feature extraction. 

\marginnote{Differential equation of learning: the magic of ``fourth order.''}
The proposed computation model arises from the minimization of an appropriate
set of constraints that is expressed by the associated cognitive action. 
The Euler-Lagrange equations of this action return a trajectory that is 
the minimum of the action under boundary conditions
that impose the vanishing of the derivates of the filter parameters.
The differential equations of the trajectory has forth-order and turns out to be the
simplest model which can simultaneously guarantee the minimization of the action
and the asymptotic stability. 

\marginnote{Blurring and day-night rhythm.}
A fundamental result
on the proposed theory concerns the structure of the companion
differential equation which carries out the blurring process as well as 
the day-night rhythm cycle. This drives the trajectory towards a 
system dynamics which guarantees that the initial conditions for the
differential equation are consistent with the boundary conditions.

\marginnote{Focus of attention.}
According to the general formulation of the theory, the optimal solution
requires to disentangle the puzzle of Euler-Lagrange equations that
take the form of integro-differential equations with non-locality in both time 
and space. While one can get rid of time non-locality in different ways, 
in order to disentangle spatial non-locality it is shown that focus of attention
is a natural solution that dramatically reduces the computation burden.

\marginnote{Addressing the ``ten questions'' on the emergence of visual perception, regardless of the nature of the agent.}
The theory opens the doors to a truly new computational
framework for computer vision, where convolutional nets can learn visual features simply
by acquiring visual data with no supervision. 
In addition to this new applicative perspective, the theory addresses the ``ten questions'' on the emergence of visual perceptual skills in both humans and machines.
Interestingly, many biological solutions of evolution are naturally explained by the proposed
framework, which suggests that they are themselves driven by information-based laws.
There a number of on-going studies that emerge from the formulation given
in this paper:

\begin{itemize}
\item  {\em Optical flow equation as another Lagrangian coordinate}\\
In the paper we have assumed that the optical low is given. 
Following the basic idea of brightness invariance~(\cite{HornAI1981})
the velocity in any pixel $x$ at time $t$ satisfies the condition 
$C(x(t),t)=c$, where $c$ is a real constant. Hence we need to satisfy the
constraint
\[
	\partial_{t} C + v_{j} \partial_{j} C = 0.
\]
This constraint can be associated with the penalty 
\[
	\int_{D} d\mu \big( \partial_{t} C + v_{j} \partial_{j} C \big)^{2}. 
\]
Once this is added to the cognitive action, we are in front of a new problem with the
additional Lagrangian coordinates $v_{j}$. As suggested in~(\cite{HornAI1981}), it is 
also opportune to introduce a regularization term so as to obey to the parsimony 
principle. 

\marginnote{The reinforcement loop for optical flow and feature computation.}
Interestingly, one can start computing the velocity field by focussing only
on the terms of the action that involves the corresponding computation. As this is 
available, it can be used for feature extraction as described in this paper. 
However, it is clear that one can benefit of a positive reinforcement of the estimations
of the optical flow and of the visual features. Once the features have been determined
on the basis of a given optical flow, we can optimize the overall cognitive action again
so as to refine its estimation. This is quite reasonable: it means that whenever, we have
a cognitive mechanisms for attaching labels to moving objects, those labels clearly 
facilitate their tracking. Basically, tracking and recognition becomes two cooperating
processes that mutually benefit by a virtuous circle of interactions.
However, it interesting to notice that the optical flow can 
be determined regardless of feature extraction with good
approximation. The described process somewhat obeys to
general principles pointed out in developmental psychology
concerning the the presence of stages in cognitive processes
taking place in children (see e.g.~\cite{Piaget1958,Piaget1961}). 


\item {\em Integration with eye movements}\\
A similar mutual benefit can be experimented when pairing the computation of the
trajectory of the focus of attention with the process of feature extraction. 
A recent study on eye movement~(\cite{DBLP:conf/nips/ZancaG17}) is in fact 
based on the variational same principle 
invoked in this paper and, therefore, a natural pairing is possible. 
In particular, instead of adopting the proposed way of modeling the 
peripheral vision, we can modify the ``curiosity field'' that only depends
on the gradient of the brightness with an appropriate enriched field that
takes into account the features extracted at $(x,t)$. 
\marginnote{The reinforcement loop of focus of attention and feature computation.}
In so doing, 
instead of modeling the local details only, this new field turns out to 
depend also on the context of the pixel and can properly model 
peripheral vision processes. 
Clearly, like for the optical flow estimation, 
there is in fact a looping process that can yield a benefit in the 
process of feature extraction.

\item {\em Learning in layered architectures}\\
The variational analysis carried out in this paper for the optimal extraction of visual features
provides yet another remarkable support to the need of hierarchical organization
in neural computation. It has been shown that only deep nets can efficiently 
discover features also in the fully unsupervised framework proposed in this paper
(see question $Q6$). It has been pointed out that different layers
extract additional information simply because of the natural receptive field structure
of the filters that is derived from the need of local equations. 
Interestingly, once we realize that shallow nets cannot afford to
solve the problem efficiently, we can start thinking of a deep net since the beginning
and propose a different formulation.

\marginnote{Feature encoding with deep nets.}
Suppose we add a layer on top of $\feat$, so as
$\Feat2 = \varphi_{2,1} \star \feat$. In this case,  we have two unknown
filters $\varphi_{1,0}$ and  $\varphi_{2,1}$. 
We can reformulate the problem of learning  $\varphi_{1,0}$ and  $\varphi_{2,1}$
by composing a new cognitive action, where we add the new constraints
$\feat = \varphi_{1,0} \star C$ and $\Feat2 = \varphi_{2,1} \star \feat$.
In so doing, in addition to the the unknown filters $\varphi_{1,0}$ and  $\varphi_{2,1}$
we also involve the variables $\Feat1$ and $\Feat2$. Of course, this 
way of enforcing constraints that express the deep net model can be 
extended to any depth. 


\item {\em Introduction of the purpose of the agent}\\
While the feature extraction process described in this paper has a truly unsupervised
nature, we can think of environmental interactions that provide fundamental 
information for the achievement of a specific task. For example, as already
pointed out, navigation in the environment and object recognition are remarkably different. 
In humans and primates this corresponds with a separate computational process
taking place in the dorsal and ventral pathways. 
Interestingly, the variational framework described in this paper need not to change
to acquire skills driven by a certain purpose. We need to compose the appropriate
cognitive action to properly model the agent environmental interaction, but the
main learning structure that emerges as Euler-Lagrange equations is the same.
In particular, the overall learning process based on the video blurring and
day-night rhythm is still at the basis of the satisfaction of the boundary conditions.
The previous remarks on the way the optical flow and also the eye movements
can improve during the acquisition of visual features indicates that, again,
the learning process must be driven by developmental phases. 
This means that the unsupervised process  described in this paper must 
be carried out until good features have been developed and, only later on,
higher level linguistic tasks must take place. Of course, this can be driven
by focussing of attention mechanisms that simply ignore complex linguistic
tasks until low level visual features have been developed.

\item {\em Statistical connections} 
Most of machine learning algorithms are developed in the framework of statistics,
so as the achievement of good generalization is strongly connected with the appropriate
selection of the samples to be used for learning. The approach to learning discussed
in this paper does not rely on a training set, but on the formalization of the life of the
agent in its own visual environment, so as there is no neat distinction between 
training and test sets. Clearly, we need to devise a different way of capturing the
consistency of the learning process that is not necessarily driven directly by statistical
principles that rely on sample complexity analyses. 
Interestingly, in the case of asymptotically stable Euler-Lagrange differential
equations and of almost periodic\cite{besicovitch1954almost} video signals
we conjecture that the formulation given in this paper allows us 
to  establish close links with classic statistical assessment. 
This is due to the minimization property of the action that has
been derived in this paper and to the causality of the agent that derives
from the blurring process.

\item {\em Energy analysis}
While the day-night rhythm alternation, along with the associated relaxation phase is 
technically sound, one promptly realizes that in order to gain visual concepts, 
the agent needs to develop a filter solution $q$ that, 
after a certain learning period, ends up
into a stable point at the beginning of every day. This is fundamental in order to 
guarantee a consistent scheme in response to visual stimulus. 
How can such a configuration
be reached? Clearly, the filters are daily updated beginning from the agent birth. 
The  given computational laws of learning, that include video blurring and
the day-night rhythm alternation indicate that the filter updating is quite robust
and proceeds with a velocity that very much depends on the monotonic function $h$,
which plays an important role in stability. We conjecture that the process of 
building the convolutional filters is in fact one which dissipates energy to construct
ordered configurations, driven by the video stream regularities. 
The general analysis given in~\cite{DBLP:journals/tcs/BettiG16} could be
used to explore this conjecture.

\end{itemize}


{\bf Acknowledgments}
Some preliminary ideas beyond this paper come from discussions 
carried out for years in the SAILab with Stefano Melacci, Marco Maggini, and Marco Lippi on the notion of developmental visual agents \url{http://dva.dii.unisi.it/}. 
We also thank Marcello Pelillo and Fabio Roli for discussions on video blurring in newborns which, later on, have stimulated its computational basis. 

\newpage
\begin{appendices}
\section{Variation of the motion term}
\label{VariationMotionTerm}
We will show in some details how the variation of the motion term $\omega_m(\varphi)=
\int dzd\tau\, f(z,\tau) (D_\tau\feat_k(z,\tau))^2$
gives Eq.~(\ref{MotionELT}).
For compactness we let $d/dt\equiv D_t$, then
\[
 \frac{\delta \omega_m(\varphi)}{\delta\varphi_{ij}(x,t)}
 =\!\!\!\int dzd\tau d\xi\, f(z,\tau) D_\tau\feat_k(z,\tau) \frac{\delta}{\delta\varphi_{ij}(x,t)}
 D_\tau(C_m(z-\xi,\tau)\varphi_{km}(\xi,\tau)).
\]
now using the property $\delta/\delta u(x)\int dy\, A(y)\dot u(y)=-\int dy\, \dot A(y)\delta(x-y)$
(valid under the assumption that in doing the variations all the boundary terms are zero) we have
\[\begin{split}
  \frac{\delta \omega_m(\varphi)}{\delta\varphi_{ij}(x,t)}
  =\int dz\bigl[&f(z,t)D_t\feat_i(z,t)D_tC_j(z-x,t)\\
    &-\del_t(f(z,t)C_j(z-x,t)D_t\feat_i(z,t))\bigr].\\
\end{split}\]
We can rewrite this variation in terms of the filters $\varphi$ using the expression for
$D_t\feat_i(z,t)=\int dy\,[\partial_t\varphi_{ij}(y,t)C_k(z-y,t)+\varphi_{ik}(y,t)D_tC_k(z-y,t)]$:
\[\begin{split}
    \int dzdy\, \Big\{&f(z,t)\big[\partial_t\varphi_{ij}(y,t)C_k(z-y,t)\\
    &\phantom{f(z,t)\big[}+\varphi_{ik}(y,t)D_tC_k(z-y,t)\big] D_tC_j(z-x,t)\\
    &-\del_t\big[f(z,t)C_j(z-x,t)\big(\partial_t\varphi_{ij}(y,t)C_k(z-y,t)\\
    &\phantom{-\del_t\big[f(z,t)C_j(z-x,t)\big(}+\varphi_{ik}(y,t)D_tC_k(z-y,t)\big)\Big\}.
  \end{split}\]
Then expanding this expression we obtain
\[\begin{split}
  \frac{\delta \omega_m(\varphi)}{\delta\varphi_{ij}(x,t)}=\!\!\!
  \int dy\, \Big\{&-\Big(\int dz\,f(z,t)C_j(z-x,t)C_k(z-y,t)\Big)\del_t^2\varphi_{ik}(y,t)\\
  &+\Big(\int dz\,f(z,t)D_tC_j(z-x,t)C_k(z-y,t)\\
  &\phantom{\Big(\int dz\,}-\del_t[f(z,t)C_j(z-x,t)C_k(z-y,t)]\Big)\del_t\varphi_{ik}(y,t)\\
  &+\Big(\int dz\,f(z,t)D_tC_j(z-x,t)D_tC_k(z-y,t)\\
  &\phantom{\Big(\int dz\,}-\del_t[f(z,t)C_j(z-x,t)D_tC_k(z-y,t)]\Big)\varphi_{ik}(y,t)\Big\},\\
\end{split}\]
which is exactly the same expression of Eq.~(\ref{MotionELT})
once we define $A$, $B$ and $C$ as in
Eq.~(\ref{MI-EL-terms}).


\section{Linearization}
\label{linearization-appendix}

Let us now define the sets
\[\begin{split}
&I=\{\,ijkl\mid i,j,k,l\in\N,\quad 0\le i<n,\,
0\le j<m,\,-\ell\le k\le\ell,\,-\ell\le l\le\ell\,\}\\
&\bar I=\{\,ijkl\mid i,j,k,l\in\N,\quad 0\le i<n,\,
0\le j<m,\,0\le k\le2\ell,\,0\le l\le2\ell\,\}\\
&\bar I^+=\{\,ijkl\mid i,j,k,l\in\N,\quad 0\le i<n,\,
0\le j<m,\,0\le k\le3\ell,\,0\le l\le3\ell\,\}\\
\end{split}
\]
together with the map $\tau_\ell$ defined by $\tau_\ell(i\,j\,k\,l)=ij(k+\ell)
(l+\ell)$. This map clearly is a bijection between $I$ and $\bar I$
with the obvious inverse $\tau_{-\ell}$.
The proper linearization is achieved through the map $r\colon \bar I^+\to J^+=\{0,\dots,N\}$
with $N=nm(3\ell+1)^2-1$
\[
  r(i\,j\,k\,l)=\left[{i,\atop n,}\,{j,\atop m,}\,{\hfill k,\atop 3\ell+1,}\,
    {\hfill l\atop 3\ell+1}\right]=l+k(3\ell+1)+j(3\ell+1)^2+i(3\ell+1)^2\cdot m.
\]
And since this is a mix-radix system with bases $3\ell+1$, $m$ and $n$
(see \cite{taocp2} page 208) $r$ is bijective and to map a number $u\in J^+$ back to $\bar I^+$ one
just have to apply the traditional division algorithm with slight modifications
(the precise algorithm is explained in exercise 2 page 327 of \cite{taocp2}).
Let us also define $J=r(\bar I)$.

Given the maps $\varphi_{ijkl}\colon\R^+\to \R$, $t\mapsto\varphi_{ijkl}(t)$ for
each $j\in J^+$ we can define the map $q_j\colon \R^+\to \R$
\[t\mapsto q_j(t):=\varphi_{\tau_{-\ell}(r^{-1}(j))}(t).\]
Similarly for the color field maps $C_{jkl}\colon \R^+\to [0\dts a]$ with $a<\infty$
defined by $t\mapsto C_{jkl}(t)$, we first extend, for every $0\le i<n$, to a three index
object $\tilde C_{ijkl}(t)=C_{jkl}(t)$ and then we define for every $j\in J^+$
the function $\Gamma_j\colon \R^+\to [0\dts a]$ so that
\[t\mapsto \Gamma_j(t):=\tilde C_{\tau_{-\ell}(r^{-1}(j))}(t).\]
With this definitions we can now rewrite $C^\sharp_{ix}$ in terms of the new
functions $q$ and $\Gamma$; we have:
\[
\begin{split}
 \feat_{ix}^\sharp(t)&=\frac{1}{n}+\sum_{k=0}^{m-1}
\sum_{\scriptstyle 0\le\xi_1\le2\ell\atop\scriptstyle 0\le\xi_2\le2\ell}
\varphi_{ik(\xi_1-\ell)(\xi_2-\ell)}(t)\tilde C_{ik(x_1-\xi_1+\ell)(x_2-\xi_2+\ell)}(t)\\
&=\frac{1}{n}+\sum_{k=0}^{m-1}
\sum_{\scriptstyle 0\le\xi_1\le2\ell\atop\scriptstyle 0\le\xi_2\le2\ell}
\varphi_{\tau_{-\ell}(ik\xi_1\xi_2)}(t)\tilde C_{\tau_{-\ell}(ik(x_1-\xi_1+2\ell)(x_2-\xi_2+2\ell)}(t).
\\\end{split}\]
Then if we now define $\gamma_x(i\,j\,k\,l):=i\,j\,(x_1-k+2\ell)\,(x_2-l+2\ell)$
from the definition of $q$ and $\Gamma$ we have that
\[ \feat_{ix}^\sharp(t)=\frac{1}{n}+\sum_{k=0}^{m-1}
\sum_{\scriptstyle 0\le\xi_1\le2\ell\atop\scriptstyle 0\le\xi_2\le2\ell}
q_{r(ik\xi_1\xi_2)}(t) \Gamma_{r(\gamma_x(ik\xi_1\xi_2))}(t).\]
Notice that $r(\gamma_x(i\,k\,\xi_1\,\xi_2))$ is well defined since $0\le x_i-\xi_i+2\ell\le3\ell$
for $i=1,2$ whenever $0\le x_i\le \ell$ and $0\le\xi_i\le2\ell$. Moreover if
we set $J\supset J_i:=\{\,j\in J\mid r^{-1}(j)=i\,k\,\alpha\,\beta,\quad
0\le j<m,\,0\le \alpha\le2\ell,\,0\le \beta\le2\ell\,\}$ for $i=0,\dots, n-1$
then we finally have
\[ \feat_{ix}^\sharp(t)=\frac{1}{n}+\sum_{j\in J_i}
q_j(t) \Gamma_{T_x(j)}(t),\]
where $T_x(j)=r(\gamma_x(r^{-1}(j)))$. Notice that for any $j\in J$ we may have
$T_x(j)\not\in J$, however we always have that $T_x(j)\in J^+$ and therefore
we may think of $T_x$ as a map between $J$ and $J^+$.

We will now show how it is possible to rewrite each term of the cognitive action
in terms of the new variables $q_j$. Before doing this let us collect the relevant
variables for the feature extraction into a vector: Let
$q:=(q_{i_1},q_{i_2},\dts,q_{i_L})'$, where $i_1<i_2<\cdots<i_L$ are the indices that
form $J$, i.e. $J=\{i_1,i_2,\dots i_L\}$ with $i_j\in J^+$, and $L=nm(2\ell+1)^2$.
This vector naturally defines a map $q\colon\R^+\to \R^L$. Moreover we will assume that
matrix multiplications and scalar products are computed summing the indices of $J$:
$(A x)_i=\sum_{j\in J} a_{ij} x_j$ and $a\cdot b =\sum_{j\in J} a_j b_j$.

Let us begin with the entropy term
\[\begin{split}
  \Bigl(\int_Dd\mu\, \feat_i\Bigr)^2&\mathrel{\mathop{\longrightarrow}^{\rm disc\,\,\,}}
  \sum_{i=0}^{n-1}\Bigl(\int_0^Tdt\, h(t)\sum_{x\in X^\sharp} g_x \bigl(\frac{1}{n}
  +\sum_{j\in J_i}q_j\Gamma_{T_x(j)}\bigr)\Bigr)^2\\
  &=\frac{1}{n}+\frac{2}{n}\int_0^Tdt\, h b\cdot q+
  \sum_{i=0}^{n-1}\Bigl(\int_0^Tdt\, h
  \sum_{j\in J_i} b_j q_j\Bigr)^2,\\
\end{split}\]
where we have set $b_j(t):=\sum_{x\in X^\sharp} g_x \Gamma_{T_x(j)}$ and
$b=(b_{i_1}, b_{i_2},\dots, b_{i_L})'$.

The conditional entropy instead gives
\[\begin{split}
    \int_Dd\mu\, \feat_i^2&\mathrel{\mathop{\longrightarrow}^{\rm disc\,\,\,}}
    \sum_{i=0}^{n-1}\int_0^Tdt\, h(t)\sum_{x\in X^\sharp} g_x \Bigl(\frac{1}{n}
    +\sum_{j\in J_i}q_j\Gamma_{T_x(j)}\Bigr)^2\\
    &=\frac{1}{n}+\frac{2}{n}\int_0^Tdt\, h b\cdot q
    +\int_0^Tdt\, h(t)\sum_{i=0}^{n-1}\sum_{j,\ell\in J_i}q_j M_{j\ell} q_\ell\\
        &=\frac{1}{n}+\frac{2}{n}\int_0^Tdt\, h b\cdot q
    +\int_0^Tdt\, h(t) q\cdot M^\natural q\\
\end{split}\]
where $M_{jl}=\sum_{x\in X^\sharp} g_x \Gamma_{T_x(j)} \Gamma_{T_x(l)}$ and
we have introduced the following notation: Given a matrix $Q$ over the indices of $J$
i.e. given the numbers $Q_{jl}$ with $j,l\in J$ we define $Q^\natural$ such that
$(Q^\sharp)_{jl}=Q_{jl}\cdot [\hbox{$j$ and $l\in J_k$ for some $k$}]$. In this way
$\sum_{i=0}^{n-1}\sum_{j,l\in J_i}q_j Q_{jl} q_l=\sum_{j,l\in J} q_j Q^\natural_{jl} q_l
=q\cdot Qq$.

The motion invariance term is already discussed in Section~\ref{discretization-section}, so we can move on
to the probabilistic constraints.
\[\begin{split}
    \int_Dd\mu\, \Bigl(\sum_{i=0}^{n-1}\feat_i^2-1\Bigr)&\mathrel{\mathop{\longrightarrow}^{\rm disc\,\,\,}}
    \int_0^Tdt\, h(t)\sum_{x\in X^\sharp} g_x \Bigl(\sum_{i=0}^{n-1}
    \sum_{j\in J_i}q_j\Gamma_{T_x(j)}\Bigr)^2\\
    &=\int_{0}^Tdt\, h(t) q\cdot Mq.\\
\end{split}\]
The sign term instead gives:
\[
    \int_Dd\mu\, \feat_i [\feat_i<0]\mathrel{\mathop{\longrightarrow}^{\rm disc\,\,\,}}
    \int_0^Tdt\, h(t) w_s(t,q)
   \]
where
\[w_s(t,q):=\sum_{i=0}^{n-1}\sum_{x\in X^\sharp} g_x
    \Bigl(\frac{1}{n} +\sum_{j\in J_i}q_j\Gamma_{T_x(j)}\Bigr)
    \Bigl[\sum_{k\in J_i}q_k\Gamma_{T_x(k)}<-\frac{1}{n}\Bigr].
\]
Therefore the action in the discretized formulation becomes
\begin{equation}
  \begin{split}
  {\cal A}(q)= &\sum_{i=0}^{n-1}\Bigl(\int_0^Tdt\, h
  \sum_{j\in J_i} b_j q_j\Bigr)^2+ \int_0^Tdt\, h \Bigl(\frac{\alpha}{2}|\ddot q|^2+
  \frac{\beta}{2}|\dot q|^2+\frac{\lambda_M}{2}\dot q\cdot M^\natural \dot q\\
  &+\lambda_M q\cdot N^\natural \dot q+\frac{1}{2}q\cdot \bigl(k-\lambda_CM^\natural+
  \lambda_1 M+\lambda_M O^\natural\bigr)q\\
  &+\frac{1-\lambda_C}{n}b\cdot q+w_s(t,q)\Bigr).\\
\end{split}
\label{appendix-action-1}
\end{equation}
Also if we approximate the entropy term as suggested in the first point at
the end of Section~\ref{Cognitive-action-section}. With this substitution the only
change is that the first non-local term in Eq.~(\ref{appendix-action-1})
that becomes
\[
  \frac{1}{2}\sum_{i=0}^{n-1}\int_0^Tdt\, h
  \Bigl(\sum_{j\in J_i} b_j q_j\Bigr)^2=  \frac{1}{2}\int_0^Tdt\, h\sum_{i=0}^{n-1}
  \sum_{j,l\in J_i}q_j b_jb_l q_l
  =\frac{1}{2}\int_0^Tdt\,q\cdot B^\natural q,
\]
where $B_{jl}=b_j b_l$. So the action becomes
\begin{equation}
  \begin{split}
  {\cal A}(q)=  \int_0^Tdt\, h \Bigl(&\frac{\alpha}{2}|\ddot q|^2+
  \frac{\beta}{2}|\dot q|^2+\frac{\lambda_M}{2}\dot q\cdot M^\natural \dot q
  +\lambda_M q\cdot N^\natural \dot q\\
  &+\frac{1}{2}q\cdot \bigl(k+B^\natural-\lambda_CM^\natural+
  \lambda_1 M+\lambda_M O^\natural\bigr)q\\
  &+\frac{1-\lambda_C}{n}b\cdot q+w_s(t,q)\Bigr).\\
\end{split}
\label{appendix-action-2}
\end{equation}
If we let $k+B^\natural-\lambda_CM^\natural+
\lambda_1 M+\lambda_M O^\natural\equiv P$, $\beta+\lambda_M M^\natural\equiv R$
and we use the convention that $\hat f=f h$, we can rewrite the action in the more
compact way
\begin{equation}
  \begin{split}
    {\cal A}(q)=  \int_0^Tdt\,\Bigl(&\frac{\hat \alpha}{2}|\ddot q|^2
  +\frac{1}{2}\dot q\cdot \hat R \dot q
  +\lambda_M q\cdot N^\natural \dot q
  +\frac{1}{2}q\cdot \hat P q\\
  &+\frac{1-\lambda_C}{n}b\cdot q+w_s(t,q)\Bigr).\\
\end{split}
\label{appendix-action-2}
\end{equation}


 \section{Variation in the discrete}
\label{discrete-variation}
Let us consider the first variation of the functional in Eq.~(\ref{action-in-the-discrete}).
Actually since the variation of the second and third line of that equation is immediate
let us just focus on the variation of the first line and define
\[{\cal A}_1(q)=\int_0^Tdt\, \Bigl(\frac{\hat\alpha}{2}|\ddot q|^2+
\frac{\hat\beta}{2}|\dot q|^2+\hat\gamma\ddot q\cdot \dot
q+\frac{\lambda_M}{2}\dot q\cdot \hat M^\natural \dot q +\lambda_M q\cdot
\hat N^\natural \dot q\Bigr)\]
Let $\psi(s)={\cal A}_1(q+s v)$ for any $v\in V$, then the first variation
in the direction of $v$ of this functional is of course $\psi'(0)$. Then:
\[\psi'(0)=\int_0^T dt\, \bigl\{(\hat\alpha\ddot q+\hat \gamma\dot q)\cdot\ddot v
  +[(\hat\beta+\lambda_M\hat M^\natural)\dot q+\hat\gamma\ddot q+\lambda_M
  (\hat N^\natural)'q]\cdot\dot v+\lambda_M\hat N^\natural \dot q\cdot v\bigr.\}
\]
Now we start integrating by parts and we repeatedly use the fact that if $v\in V$,
then $v(0)=\dot v(0)=0$, then we get
\[
  \begin{split}
\psi'(0)=&\big[(\hat\alpha\ddot q+\hat \gamma\dot q)\dot v+
\big((\hat\beta+\lambda_M\hat M^\natural)\dot q+\hat\gamma\ddot
q+\lambda_M (\hat N^\natural)'q -(\hat\alpha\ddot q+\hat \gamma\dot q)\dot{ }\big)v\big]_{t=T}
\\
&+\int_0^T \big\{(\hat\alpha\ddot q+\hat \gamma\dot q)\,\ddot{ }\,
- \big((\hat\beta+\lambda_M\hat M^\natural)\dot q+\hat\gamma\ddot
q+\lambda_M (\hat N^\natural)'q\big)\dot{ }\, +\lambda_M\hat N^\natural \dot q\big\}\cdot v
\end{split}
\]
The E-L equations are obtained when we impose $\psi'(0)=0$ and since we can always chose a
variation such that $v(T)=\dot v(T)=0$ using the fundamental lemma of the calculus of
variations
\begin{equation}\begin{split}
  \hat\alpha q^{(4)}+2\dot{\hat\alpha} q^{(3)}&+(\ddot{\hat \alpha}+\dot{\hat \gamma}-\hat \beta
  -\lambda_M\hat M^\natural)\ddot q\\
  &+(\ddot {\hat \gamma}-\dot{\hat\beta}-\lambda_M
  (\dot{\hat M}^\natural+(\hat N^\natural)'-\hat N^\natural))\dot q-\lambda_M
  (\dot{\hat N}^\natural)'q=0.\end{split}
\label{appendix-e-l-eq}
\end{equation}
Putting back the E-L equations into the expression of the variation we also obtain the
appropriate boundary conditions
\[
\begin{split}
  &\hat\alpha\ddot q(T)+\hat\gamma\dot q(T)=0;\\
  &\hat\alpha q^{(3)}(T)-\dot{\hat\alpha}\ddot q(T)+(\hat\beta+\lambda_M\hat M^\natural -\dot{\hat\gamma})
    \dot q(T)+\lambda_M(\hat N^\sharp)'q(T)=0.
  \end{split}
\]
Notice that when the input is zero at $t=T$, then the second of this conditions
reduces to $\hat\alpha q^{(3)}(T)-\dot{\hat\alpha}\ddot q(T)+(\hat\beta-\dot{\hat\gamma}) \dot q(T)=0$ which  does not involve the value of $q$ at $T$ but only the values of its derivatives.

In order to complete the variation of the functional in Eq.~(\ref{action-in-the-discrete})
we have to add to Eq.~(\ref{appendix-e-l-eq}) the term $\bigl(k \hat I+\hat B^\natural
-\lambda_C\hat M^\natural+
\lambda_1 \hat M+\lambda_M \hat O^\natural\bigr)q+\frac{1-\lambda_C}{n}\hat b+\nabla_q
\hat w_s(t,q)$. This means
that, referring to Eq.~(\ref{ForthOrderDiff}) the coefficients has the following form:
\begin{subequations}\begin{align}
    &A(t)=\frac{\dot h}{h};\\
    &B(t)=\frac{\ddot h}{h}+\frac{\gamma\dot h}{\alpha h}
    -\frac{\beta}{\alpha}-\frac{\lambda_M}{\alpha}M^\natural;\\
    &C(t)=\frac{\gamma \ddot h}{\alpha h}-\frac{\beta\dot h}{\alpha h}
    -\frac{\lambda_M\dot h}{\alpha h}-\frac{\lambda_M}{\alpha}(\dot M^\natural+(N^\natural)'
    -N^\natural);\\
    &D(t)=-\frac{\lambda_M \dot h}{\alpha h} (N^\natural)'+\frac{\lambda_M}{\alpha}(O^\natural
    -(\dot N^\natural)')+\frac{k}{\alpha}+\frac{B^\natural}{\alpha}-\frac{\lambda_C}{\alpha}
    M^\natural +\frac{\lambda_1}{\alpha}M;\\
    &F(t,q)=\frac{1-\lambda_C}{\alpha n}b+\frac{1}{\alpha}\nabla_q w_s(t,q).
\end{align}
\end{subequations}
The matrices $M$, $N$ and $O$ are defined in Eq.~\ref{MI-matrices},
the vector $b$ and the function $w_s$ are instead introduced in
Eq.~\ref{sign-and-b-term} and $B_{jl}=b_j b_l$. 

\end{appendices}

\bibliography{corr,nn}
\bibliographystyle{plain}
\end{document}